%% file: manuscript.tex
\documentclass{article}
\usepackage{arxiv}

\usepackage[utf8]{inputenc} 
\usepackage[T1]{fontenc}    
\usepackage{hyperref}       
\usepackage{url}            
\usepackage{booktabs}       
\usepackage{amsfonts}       
\usepackage{nicefrac}       
\usepackage{microtype}      
\usepackage{lipsum}		
\usepackage{graphicx}
\usepackage{natbib}
\usepackage{doi}
\usepackage{wrapfig}
\usepackage{multirow}
\usepackage{amsmath}
\usepackage{multirow}
\usepackage{bbm}
\usepackage{wrapfig}
\usepackage[font=small,skip=0pt]{caption}
\usepackage{soul}
\usepackage{lineno}
\usepackage[normalem]{ulem}

\title{Self-Supervised Learning for Videos: A Survey}

\author{ \href{https://orcid.org/0000-0001-7350-5008}{\includegraphics[scale=0.06]{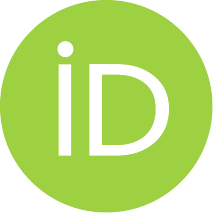}\hspace{1mm}Madeline C.~Schiappa}
		\\
	Center for Research in Computer Vision\\
	University of Central Florida\\
	Orlando, FL 32816 \\
	\texttt{madelineschiappa@knights.ucf.edu} \\
	\And
	\href{https://orcid.org/0000-0003-4052-6798}{\includegraphics[scale=0.06]{orcid.pdf}
	\hspace{1mm}Yogesh S.~Rawat} \\
	Center for Research in Computer Vision\\
	University of Central Florida\\
	Orlando, FL 32816 \\
	\texttt{yogesh@crcv.ucf.edu} \\
	\And
    \href{https://orcid.org/0000-0001-6172-5572}{\includegraphics[scale=0.06]{orcid.pdf}
	\hspace{1mm}Mubarak Shah} \\
	Center for Research in Computer Vision\\
	University of Central Florida\\
	Orlando, FL 32816 \\
	\texttt{shah@crcv.ucf.edu} \\
}

\renewcommand{\headeright}{}
\renewcommand{\undertitle}{\vspace{-5ex}}
\renewcommand{\shorttitle}{Self-Supervised Learning for Videos: A Survey}

\hypersetup{
pdftitle={Self-Supervised Learning for Videos: A Survey},
pdfsubject={cs.CV},
pdfauthor={Madeline C.~Schiappa, Yogesh S.~Rawat, Mubarak Shah},
pdfkeywords={robustness, video robustness, text robustness, text-to-video retreival, video retreival, video-language modeling, computer vision},
}
\begin{document}
\maketitle





\begin{abstract}

The remarkable success of deep learning in various domains relies on the availability of large-scale annotated datasets. 
However, obtaining annotations is expensive and requires great effort, which is especially challenging for videos. 
Moreover, the use of human-generated annotations leads to models with biased learning and poor domain generalization and robustness.
As an alternative, self-supervised learning provides a way for representation learning which does not require annotations and has shown promise in both image and video domains. Different from the image domain, learning video representations are more challenging due to the temporal dimension, bringing in motion and other environmental dynamics. This also provides opportunities for video-exclusive ideas that advance self-supervised learning in the video and multimodal domain. In this survey, we provide a review of existing approaches on self-supervised learning focusing on the video domain. We summarize these methods into four different categories based on their learning objectives: 1) \textit{pretext tasks}, 2) \textit{generative learning}, 3) \textit{contrastive learning}, and 4) \textit{cross-modal agreement}. We further introduce the commonly used datasets, downstream evaluation tasks, insights into the limitations of existing works, and the potential future directions in this area. \footnote{GitHub Project Link: \url{https://github.com/Maddy12/SSL4VideoSurvey}}

\end{abstract}



\maketitle

\section{Introduction}
\label{sec:intro}
\input{sections/introduction}

\section{Preliminaries}
\label{sec:prelim}
\input{sections/preliminaries}

\section{Self-Supervised Learning Approaches}
\label{sec:body}
We split the works in video self-supervised learning into four high level categories: pretext, generative, contrastive, and cross-modal agreement. We will discuss each in the following subsections.

\subsection{Pretext Learning}
\label{sec:pretext}
\input{sections/pretext}

\subsection{Generative Approaches}
\label{sec:generative}
\input{sections/generative}

\subsection{Contrastive Learning}
\label{sec:contrastive}

\input{sections/contrastive}

\subsection{Cross-Modal Agreement}
\label{sec:crossmodal}
\input{sections/crossmodal}

\section{Summary and Future Directions}
\label{sec:discussion}
\input{sections/summary_future_work}
\bibliographystyle{unsrtnat}
\bibliography{acmart}


\end{document}

%% file: sections/introduction.tex
The requirement of labelled samples at scale limits the use of deep networks for problems where data is limited and obtaining annotations is not trivial, such as medical imaging \cite{dargan2020survey} and biometrics \cite{biometric_survey}.
While pre-training on a large-scale labelled dataset such as ImageNet \cite{krizhevsky2012imagenet} and Kinetics \cite{kay2017kinetics} does improve performance, 
there are several limitations and drawbacks in this approach. These drawbacks include annotation cost \cite{yang2017suggestive, cai2021revisiting}, presence of annotation bias \cite{chen2021understanding, rodrigues2018deep}, lack of domain generalization \cite{wang2021generalizing, hu2020strategies, Kim_2021_ICCV}, and lack of robustness \cite{hendrycks2019benchmarking,hendrycks2021many}.

\begin{figure}[t!]
    \centering
    \includegraphics[width=.80\linewidth]{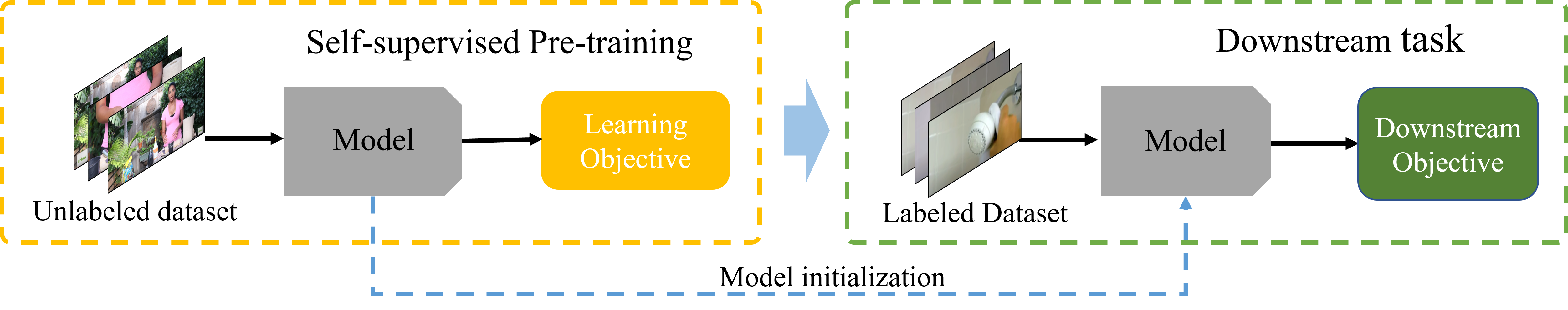}
    \caption{An illustration of using a pre-trained model trained using self-supervised learning for a downstream task. The process starts by pre-training a model on an unlabeled dataset using a self-supervised learning objective. Once trained, the learned weights are used as model initialization for a smaller, labelled dataset on a downstream task
    }
    \label{fig:transfer_learning}
\end{figure}

\begin{figure}
    \centering
    \includegraphics[width=\linewidth]{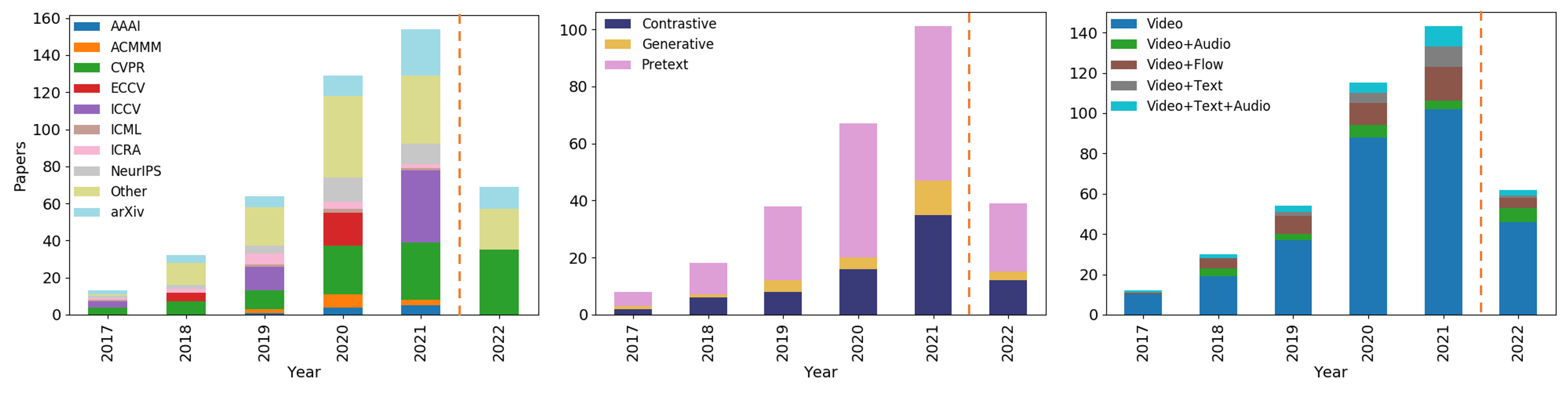}
    \caption{Statistics of self-supervised (SSL) video representation learning research in recent years.
    From left to right we show a) the total number of SSL related papers published in top conference venues, b) categorical breakdown of the main research topics studied in SSL, and (c) modality breakdown of the main modalities used in SSL. The year 2022 remains incomplete because a majority of the conferences occur later in the year.}
    \label{fig:fancy_plots}
\end{figure}

Self-Supervised learning (SSL) has emerged as a successful way for pre-training deep models to overcome some of these issues. It is a promising alternative where a model can be trained on large-scale datasets without the need of labels \cite{jing2020self} and with improved generalizability. SSL trains the model using a learning objective derived from the training samples itself. Typically, the pre-trained model is then finetuned on the target dataset as shown in Figure \ref{fig:transfer_learning}. 

SSL has been found effective for different domains such as images \cite{radford2021learning, gidaris2018unsupervised, doersch2015unsupervised, noroozi2016unsupervised, Wei_2019_CVPR, kim2018learning}, natural language \cite{radford2021learning, devlin2018bert, vaswani2017attention}, graphs \cite{liu2021graph}, point-clouds \cite{sharma2020self, huang2021spatio, guo2020deep}, etc. Motivated by the success in the image domain, there has been great interest in adopting SSL for the video domain. However, it is not trivial to extend the image-based approaches directly to video due to time. Some early works in this area have focused on extending works in the image-domain for videos without consideration for time \cite{jing2019selfsupervised, tao2020selfsupervised, Wang_2019_CVPR_pretext, noroozi2016unsupervised, ahsan2018video, ijcai2021, kim2019selfsupervised}. Some recent approaches focused on temporal signal where the learned representations also capture the temporal dimension of videos \cite{Yao_2020_CVPR, wangjianglie2020, jenni2020video, Benaim_2020_CVPR, 9412071, 10.1007/978-3-319-46448-0_32, Lee_2017_ICCV, Fernando_2017_CVPR, Xu_2019_CVPR}. While not trivial, work in this area has more attention of researchers as shown in Figure \ref{fig:fancy_plots}. 
This figure shows a breakdown of the number of papers published on ``self-supervised video learning'' in top conference venues, research category, and modalities. There has been a significant increase in the number of papers and it is expected to continue to do so in future years. This survey aims to summarize the representative recent works in this area. 


This increase in interest from multiple domains results in a wide variety of approaches all trying to solve this important issue of training without large-scale, expensively annotated datasets. There are some recent surveys on SSL for images \cite{technologies9010002, jing2020self}. In \cite{technologies9010002}, the authors discussed contrastive self-supervised learning focusing on the image domain. Multiple categories were covered in \cite{jing2020self} including generation-based, context-based, free-semantic labeled base, and cross-modal based (i.e. optical flow), also focused on image domain. While both these works briefly cover video, they focus more on the image domain and do not comprehensively discuss the multimodal aspect.
Similarly in \cite{liu2021self}, the authors focus on a wider scope which covers computer vision, natural language processing, and graph learning. In contrast to these existing surveys, our focus is very specific to the video domain with the inclusion of multimodal approaches that have not been covered in previous surveys.

\begin{figure}
    \centering
    \includegraphics[width=.70\linewidth]{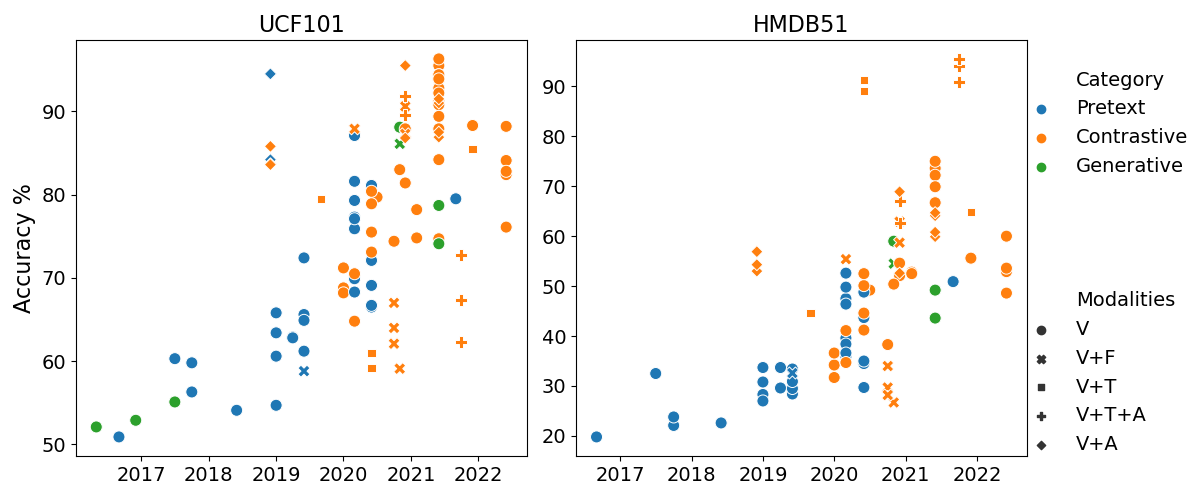}
    \caption{
    Action recognition performance of models over time for different self-supervised strategies including different modalities: video-only (V), video-text (V+T), video-audio (V+A), video-text-audio (V+T+A). More recently, contrastive learning has become the most popular strategy.
    }
    \label{fig:action_recognition_scatter_plot}
\end{figure}

Multimodal learning is an important research area to cover because it is increasingly demonstrating improved performances for zero-shot learning \cite{Miech2020, xu-etal-2021-videoclip}. 
When we refer to multimodal or cross-modal learning, we are referring to learning using video, audio and/or text. These approaches generalize to a greater variety of visual tasks including action step localization, action segmentation, text-to-video retrieval, video captioning, etc \cite{Miech2019, Miech2020, xu-etal-2021-videoclip, Luo2020, Xu2020}. These multimodal approaches can expand to other single-modality tasks such as classification of audio events \cite{piczak2015esc}. As shown in Figure \ref{fig:action_recognition_scatter_plot}, the increase in performance of multimodal SSL has led to more approaches and more combinations of approaches that need to be organized in a comprehensive way. With the increase in utilizing text as a modality, contrastive learning has also appeared more and more in the SSL research landscape (see Figure \ref{fig:action_recognition_scatter_plot}). While \cite{technologies9010002} does touch on video-based contrastive learning, our survey expands on the recent approaches that includes the use of multiple modalities for contrastive learning. With this in mind, we cover works in multimodal areas and present a summary of results for additional tasks that these models can be evaluated on. 

In summary, this survey categorizes work into four general learning categories: pre-text, generative, contrastive and cross-modal agreement. Within these categories, some works make use of additional modalities, such as text, to develop better learning objectives. We make the following contributions in this survey,
\begin{itemize}
    \item We provide a comprehensive review of recent works on self-supervised learning in the video domain, including multimodal approaches. 
    \item We present a categorization of existing works into four groups; 1) pretext learning, 2) generative approaches, 3) contrastive learning, and 4) cross-modal agreement. We believe this will guide future research in this area.
    
    \item We summarize the tasks used for evaluation in this domain, corresponding datasets, and evaluation metrics along with performance comparison. This will be useful for benchmarking future works. 
    
    \item Finally, we discuss the limitations, open problems and future directions in this research area.
\end{itemize}

\begin{figure}
    \centering
    \includegraphics[width=\linewidth]{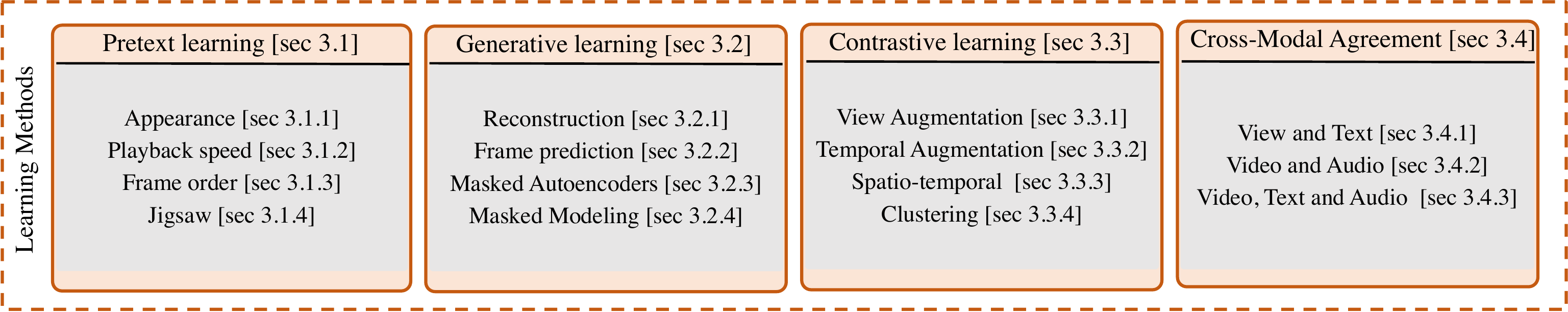}
    \caption{Works in this survey can be categorized by their self-supervised learning (SSL) method: Pretext, Generative, Contrastive and Cross-Modal agreement. 
    }
    \label{fig:survery_structure}
\end{figure}

This survey is organized as follows: Section \ref{sec:preliminaries} will go over the preliminary knowledge of self-supervised learning for videos including terminology, backbone architectures used in pre-training, datasets, and downstream tasks. Section \ref{sec:pretext} will go over works that use a pretext task. Section \ref{sec:generative} will go over works that use generative approaches. Section \ref{sec:contrastive} will go over works that use different methods to generate positive and negative pairs for contrastive learning. Section \ref{sec:crossmodal} will go over learning based on agreement between signals from video, text and/or audio. The breakdown of approaches for these categories are shown in Figure \ref{fig:survery_structure}. Finally, we will conclude with a discussion on the limitations, some open problems, and future directions in Section \ref{sec:discussion} and Section \ref{sec:conclusion}.

%% file: sections/preliminaries.tex
\label{sec:preliminaries}

We first introduce the building blocks of self-supervised learning in videos. This includes various downstream tasks in section \ref{sec:downstream}, the backbone network architectures used for training in section \ref{sec:building_blocks}, and the datasets widely used in this research in section \ref{sec:datasets}. 

\subsection{Downstream Evaluation}
\label{sec:downstream}
In order to evaluate learned video representations from pre-training, a target downstream task is solved by either fine-tuning, linear probing, or zero-shot learning. In zero-shot learning, the learned representations are used without any training on the target dataset; whereas fine-tuning and linear-probing involves additional training. We focus on the following downstream tasks for evaluation:

\textbf{Action Recognition} is the most common evaluation task for video only self-supervised learning. It is a classification problem where the goal is to classify each input video into one of the target action categories. Typically, linear layers are added to the pre-trained model to predict the action class of the input video. The model's performance for this task is measured using classification metrics: \textit{accuracy}, \textit{precision}, and \textit{recall}. 

\textbf{Temporal Action Segmentation} is a task that detects the start and end time of actions in a video. This task may also be referred to as \textit{action segmentation}. Performance is typically measured using \textit{frame accuracy} (FA), a metric that calculates the model's accuracy when predicting actions frame-by-frame.

\textbf{Temporal Action Step Localization} aims to detect which segments in a video belong to which steps in a complex activity. For each frame, a recall metric is used to define the number of step assignments that fall into the ground truth time interval, divided by the total number of steps in the video \cite{zhukov2019cross,Miech2019}. This task is used with instructional videos where a complex task, such as ``how to change a tire'', is broken down into action steps, such as ``loosen the lug nuts'' and ``replace lug nuts''.

\textbf{Video Retrieval} aims to find videos that are similar or near-duplicates of a given video. The evaluation protocol includes a gallery of videos and a set of videos with which to query. The performance is measured using standard retrieval metrics including \textit{retrieval}, \textit{recall} and \textit{similarity measures}. Recall is the fraction of retrieved videos that are relevant to the queried video. This is done at different top-k levels; typically top 1 (R@1), top 5 (R@5) and top 50 (R@50) are reported. 

\textbf{Text-to-Video Retrieval} takes a text query as input and returns corresponding matching video(s). This is commonly used when there are joint embeddings of text and video in the pre-trained representations. The metrics used are the same as that for Video-Retrieval. The key difference is the query is a textual description rather than a video. This task is often used in zero-shot learning to evaluate the quality of learned representations between text and video.

\textbf{Video Captioning} is a generative task where the input is a video clip and the output is a caption. The goal being to generate a textual caption that accurately describes the contents of the video. Accuracy is measured using a combination of \textit{BLEU} \cite{papineni2002bleu}, \textit{METEOR} \cite{banerjee2005meteor}, \textit{Rouge-L} \cite{lin2004rouge} and \textit{CIDEr} \cite{vedantam2015cider} metrics. BLEU is a precision based metric that calculates the fraction of tokens from the prediction that appear in the ground-truth. It also uses a brevity penalty that penalizes words that appear in the prediction more times than it appears in any of the ground-truth captions. Rouge-L is similar, but is a recall based metric that measures the longest common subsequence (LCS) between predicted and ground-truth. METEOR calculates a one-to-one mapping of words in predicted and ground-truth captions. Finally, Consensus-based Image Description Evaluation (CIDEr) calculates how well a predicted caption matches the consensus of a set of ground-truth captions.

\subsection{Building Blocks}
\label{sec:building_blocks}
Next, we will cover the backbone network architectures used to encode videos, text, and audio. 

\subsubsection{Visual Backbone Architectures}
\label{sec:backbone}

There are a wide range of network architectures which are used to learn video representations. These are often deep convolutional neural networks (CNNs) that use either 2D convolutions, 3D convolutions, or a combination of both. 2D CNNs encode videos at the "frame-level" where each frame is treated as an image. 3D CNNs encode videos with the dimension of time, running convolutions over both time and space. ResNet \cite{hara2018can} is the most popular architecture for both 2D and 3D CNNs. 
Other popular 3D CNN networks for video include I3D \cite{carreira2018quo}, C3D \cite{tran2015learning} and S3D \cite{xie2018rethinking}, all of which use a 3D inception module. 
Another popular architecture is the R(2+1)D \cite{Tran_2018_CVPR} 
; separating traditional 3D convolutions into a spatial 2D convolutions and a temporal 1D convolution. To improve R(2+1)D, the architecture R(2+3)D \cite{feichtenhofer2018learned, han2019video, Han2020} modify 
the last two blocks so they have 3D kernels. We have seen an increase in the use of video encoding architectures based on transformers, such as TimeSformer \cite{bertasius2021spacetime}, MViT \cite{fan2021multiscale}, ViViT \cite{arnab2021vivit}, Swin \cite{liu2021swin}. 

\subsubsection{Audio Backbone Architectures}
\begin{figure*}
    \centering
    \includegraphics[width=.55\linewidth]{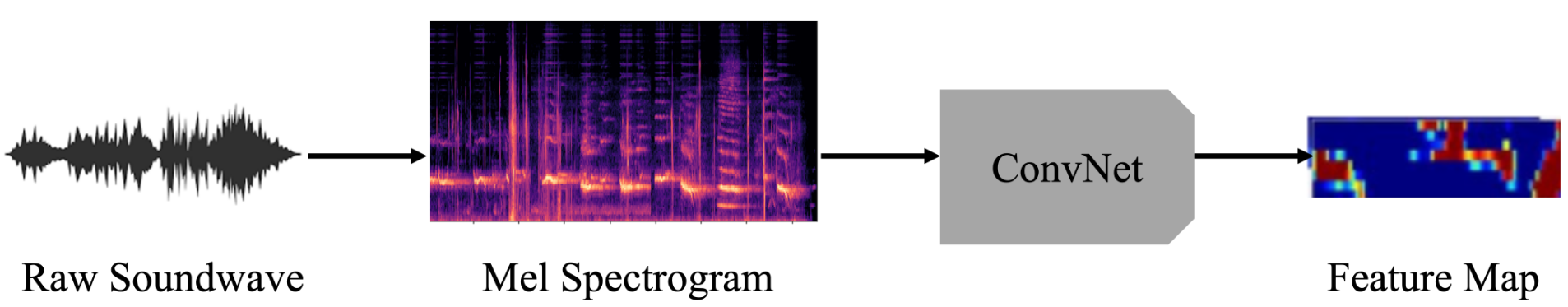}
    \caption{Audio backbone encoders convert audio waveforms to Mel Spectrograms and are typically 2D CNN-based architectures.}
    \label{fig:audio}
\end{figure*}

Audio is encoded by first converting it into spectograms (Figure \ref{fig:audio}). This representation of audio signal is generated by splitting the signal into bins. Fourier Transform is then applied to each bin to determine their respective frequencies. The most popular type of spectrogram in deep learning is the Mel Spectrogram, replacing frequency with the mel scale. CNN Models with a 2D or 3D architecture, such as ResNet \cite{he2015deep}, are then used to extract audio features from these Mel Spectograms. More details on the audio-visual deep learning can be found in \cite{zhu2021deep_audiovisual_survey}. 
There are several backbone approaches discussed in this survey \cite{zhu2021deep_audiovisual_survey}, and most of them rely on 2D or 3D ResNets to extract audio features.

\subsubsection{Text Backbone Architectures}

\label{sec:text_backbone}

\begin{wrapfigure}[18]{R}{0.45\textwidth}
\centering
    \includegraphics[width=.45\textwidth]{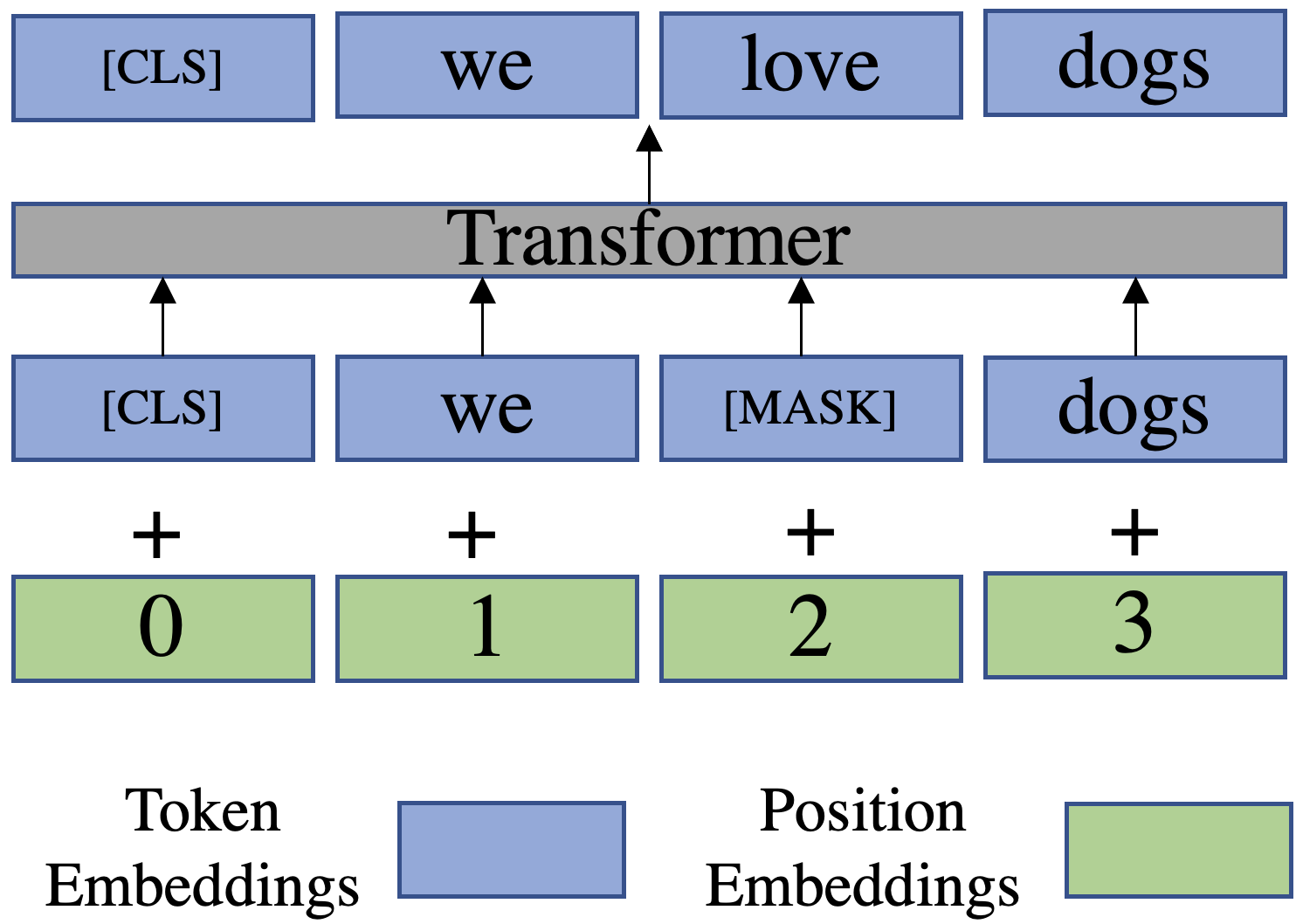}
  
    \caption{The Masked Language Modeling \cite{taylor1953cloze} approach using transformers \cite{vaswani2017attention} where each word is encoded and the transformer uses self-attention to decode each encoding to a predicted word, filling in the blanks of masked words.}
    \label{fig:mlm}
\end{wrapfigure}

The two most common approaches for text encoding are either a Word2Vec approach \cite{mikolov2013efficient} or a Bidirectional Encoder Representations from Transformers (BERT) approach \cite{devlin2018bert}. Word2Vec is a skip-gram model that uses a central word to predict n-words before and after its occurrence. BERT uses a self-attention mechanism \cite{vaswani2017attention} trained using either a Masked Language Model (MLM) \cite{taylor1953cloze} or Next Sentence Prediction (NSP) task to train. In the MLM task, shown in Figure \ref{fig:mlm}, a percentage of the word tokens are masked and the model predicts the masked tokens. NSP pair sentences with either their predecessors or ones chosen at random, and the model must predict whether they are a ``good" pair or not. 

\subsection{Datasets} 
\label{sec:datasets}
In self-supervised learning for videos, we have a large-scale dataset which is used in pre-training to learn representations. We have several smaller ``down-stream" datasets that have annotations for specific tasks. Table \ref{tab:video_datasets} summarizes the video-based datasets used in self-supervised learning. The most commonly used datasets for pre-training are the Kinetics \cite{chen2020vggsound} and HowTo100M \cite{Miech2019} datasets. The most common datasets used for downstream evaluation are UCF101 \cite{soomro2012ucf101} and HMDB51 \cite{kuehne2011hmdb}. Because there have been recent studies demonstrating that these two datasets do not necessarily require temporal learning \cite{huang2018makes}, the Something-Something (SS) \cite{goyal2017something} dataset has been proposed, but is not widely adopted for evaluation. 

\begin{table}[]
    \centering
    \caption{\textbf{Video} datasets for self-supervised training and/or evaluation and the respective tasks associated with the annotations provided.}
    \small
    \resizebox{\textwidth}{!}{\begin{tabular}{|c|c|c|c|c|c|}
        \hline
        \textbf{Dataset} & \textbf{Labels} & \textbf{Modalities} & \textbf{Classes} & \textbf{Videos} & \textbf{Tasks}  \\ 
        
        \hline
        \multirow{3}{*}{ActivityNet (ActN) \cite{Heilbron_2015_CVPR}} & Activity& Video & \multirow{3}{*}{200} & \multirow{3}{*}{19,995} & Action-Recognition \\
         &   Captions  & \multirow{2}{*}{Video+Text} &    &   & Video Captioning \cite{krishna2017dense} \\
         &    Bounding Box   &  & &      & Video Grounding \cite{zhou2019grounded} \\
        \hline
         \multirow{2}{*}{AVA \cite{gu2018ava,li2020ava}} & Activity & Video & \multirow{2}{*}{80} & \multirow{2}{*}{430} & Action-Recognition \\
         &   Face Tracks   & Video+Audio           & &       & Audio-Visual Grounding \cite{roth2020ava} \\

       \hline
        \multirow{2}{*}{Breakfast \cite{Kuehne_2014_CVPR}} & \multirow{2}{*}{Activity} & \multirow{2}{*}{Video}  & \multirow{2}{*}{10} & \multirow{2}{*}{1,989} & Action Recognition \\
                  &           &      & &       & Action Segmentation \\
    
        \hline
        \multirow{4}{*}{Charades \cite{sigurdsson2016hollywood}} & Activity & \multirow{4}{*}{Video} & \multirow{4}{*}{157} & \multirow{4}{*}{9,848} & Action-Recognition \\
        &        Objects     &      &         &    & Object Recognition \\
        &        Indoor Scenes &    &         &    & Scene Recognition \\
        &        Verbs     &        &         &    & Temporal Action Step Localization \\
        
        \hline
        \multirow{3}{*}{COIN \cite{tang2019coin}} &  Activity & \multirow{2}{*}{Video}  &  & \multirow{3}{*}{11,827} & Action-Recognition \\
        & Temporal Actions & & 180 & & Action Segmentation \\
        & ASR & Video+Text & & & Video-Retrieval \\
        
        \hline
        \multirow{2}{*}{CrossTask \cite{Zhukov_2019_CVPR}} &  Temporal Steps & \multirow{2}{*}{Video} & \multirow{2}{*}{83} & \multirow{2}{*}{4,700} & Temporal Action Step Localization \\
        & Activity & & & & Recognition \\
        
        \hline
        \multirow{2}{*}{HMDB51 \cite{kuehne2011hmdb}} & \multirow{2}{*}{Activity} & \multirow{2}{*}{Video}  & \multirow{2}{*}{51} & \multirow{2}{*}{6,849} & Action-Recognition \\
        & & & & & Video-Retrieval \\
        
        \hline
        \multirow{2}{*}{HowTo100M (How2) \cite{Miech2019}} & \multirow{2}{*}{ASR} & \multirow{2}{*}{Video+Text} & \multirow{2}{*}{-} & \multirow{2}{*}{136M} & Text-to-Video Retrieval \\
        & & & & & VideoQA \cite{li2021value} \\

        \hline
         Kinetics \cite{kay2017kinetics} & Activity & Video  & 400/600/700 & \textonehalf M & Action-Recognition \\
         
         \hline
        \multirow{4}{*}{MSRVTT \cite{Xu_2016_CVPR}} & &\multirow{4}{*}{Video+Text} & \multirow{4}{*}{20} & \multirow{4}{*}{10,000} & Action-Recognition \\
        & Activity & & & & Video-Captioning \\
        & Captions & & & & Video-Retrieval \\
        & & & & & Visual-Question Answering \\
        
        \hline
        \multirow{2}{*}{MultiThumos \cite{yeung2017moment}} & Activity & \multirow{2}{*}{Video}      & \multirow{2}{*}{65} & \multirow{2}{*}{400} & Action Recognition \\
               &        Temporal Steps & &   &       & Temporal Action Step Localization \\
        
         \hline
        \multirow{2}{*}{UCF101 \cite{soomro2012ucf101}} & \multirow{2}{*}{Activity} & \multirow{2}{*}{Video} & \multirow{2}{*}{101} & \multirow{2}{*}{13,320}  & Recognition \\
        & & & & & Video-Retrieval \\

        \hline
        \multirow{2}{*}{YouCook2 \cite{zhou2017automatic}} & \multirow{2}{*}{Captions} & \multirow{2}{*}{Video+Text} & \multirow{2}{*}{89} &  \multirow{2}{*}{2,000} & Video Captioning \\
        & & & & & Video-Retrieval \\
        
        \hline
        YouTube-8M \cite{abuelhaija2016youtube8m} & Activity & Video & 4,716 & 8M & Action Recognition \\
        \hline
        
    \end{tabular}}
    \label{tab:video_datasets}
\end{table}

\subsubsection{Video-Based Datasets}

\paragraph{ActivityNet}
This dataset contains 19,995 untrimmed user-generated videos from YouTube covering 200 different types of activities \cite{Heilbron_2015_CVPR}. Each video is annotated with temporal boundaries and has on average 1.41 activities. 

\paragraph{AVA} This dataset \cite{gu2018ava,li2020ava,roth2020ava} focuses on atomic visual actions and contains 430 videos. This dataset has three variations in total. AVA-Actions are densely annotated videos with 80 atomic visual actions localized in space in time totaling 430 15-minute movie clips.  AVA-Kinetics, which is a crossover between AVA and Kinetics with localized action labels. AVA-ActiveSpeaker is an audio-visual grounding dataset that temporally labels face tracks in video. Each face instance is labeled as speaking or not and the corresponding audio. There are a total of 3.65 million human labeled frames.  

\paragraph{Breakfast} 
This dataset consists of 1,989 filmed videos made of 10 actions related to breakfast preparation, performed by 52 different individuals in 18 different kitchens. All videos were down-sampled to a resolution of $320 \times 240$ pixels with a frame rate of 15 fps. 

\paragraph{Charades}
This dataset is a collection of 9,848 videos of daily indoor activities with an average length of 30 seconds with a total of 157 action classes \cite{sigurdsson2016hollywood}. Videos were generated by users recording themselves acting out sentences which were presented to them. These sentences contained objects and actions from a fixed vocabulary. These videos involve interactions with 46 different object classes, in 15 types of different indoor scenes and a vocabulary of 30 verbs. This dataset contains 66,500 temporal annotations for 157 action classes, 41,104 labels for 46 object classes, and 27,847 textual descriptions of the videos.

\paragraph{COIN}
This dataset is a small-scale dataset of instructional videos \cite{tang2019coin}. This dataset is a collection of 11,827 videos of 180 different tasks in 12 domains such as vehicles and gadgets. These videos are annotated with a task and temporal boundaries, allowing for models to be evaluated on action-localization. The average length of these videos is 2.36 minutes where each annotated segment lasts an average of 14.91 seconds with a total of 46,354 annotated segments.

\paragraph{CrossTask}
A dataset that consists of 4,700 YouTube videos of 83 tasks \cite{Zhukov_2019_CVPR}. Each task has a set of steps with manual description, allowing models to be evaluated on action localization of steps that make up a task. The dataset has 18 primary tasks and 65 related tasks. Videos associated with the primary tasks are collected manually and are annotated for temporal step boundaries while the videos associated with related tasks are not annotated.

\paragraph{HMDB51}
This dataset contains videos collected from a variety of sources, including commercial movies and public video hosting services \cite{kuehne2011hmdb}. There are 6,849 video clips in total averaging 10 seconds each with 51 action categories. Each action category contains at least 101 clips. The action categories are split into five types: facial actions (e.g. smiling), face-to-object interaction (e.g. eating), general body movement, body-to-object interactions (e.g. brush hair) and human-to-human interaction (e.g. hugging). 

\paragraph{HowTo100m}
One of the largest collections of user-generated instructional videos, HowTo100M (How2) \cite{Miech2019} is a collection of 136 million video clips with captions sourced from 1.2 million YouTube videos. These videos are categorized into 23,000 activities from domains such as cooking, hand crafting, personal care, gardening or fitness. The associated text is collected from subtitles downloaded from YouTube which are either user-uploaded or automatically generated. These videos are untrimmed and are often minutes-long. Some of the other dataset videos, such as those in COIN \cite{tang2019coin}, are part of this collection but lack the more detailed annotation. 

\paragraph{Kinetics} 
The most popular pre-training dataset for video-only self-supervised learning approaches \cite{kay2017kinetics}. It is a large-scale, human-action dataset containing 650,000 video clips covering either 400, 600, or 700 human action classes. Each video clip lasts around 10 seconds and is labeled with a single action class.

\paragraph{MSRVTT}
A dataset that consists of 10,000 video clips from 20 action categories \cite{Xu_2016_CVPR}. Each video clip is annotated with 20 English sentences with a total of 200,000 clip-sentence pairs. The duration of each clip is between 10 and 30 seconds. This dataset is often used for video-captioning and retrieval tasks. 

\paragraph{MultiTHUMOS}
This dataset consists of 400 untrimmed videos collected from YouTube with 65 action classes. Each video has dense, multi-label, frame-level action annotations with a total of 38,690 annotations with an average of 1.5 labels per frame and 10.5 action classes per video. 

\paragraph{Something-Something}
This dataset focuses on evaluating temporal elements in videos using everyday human activities. It is a large-scale dataset with 220,847 videos with a total of 174 activities. Some examples are  ``putting something into something'', ``turning something upside down'' and ``covering something with something''.


\paragraph{UCF-101}
One of the most popular benchmarks to evaluate models on action-recognition because of its smaller size \cite{soomro2012ucf101}. It has only 13,320 video clips which are classified into 101 categories. The categories fall within five types of activity: body motion, human-to-human interaction, human-to-object interaction, playing musical instruments and sports. The videos are user-generated, collected from YouTube, and have a fixed frame rate of 25 FPS and fixed resolution of $320 \times 240$.

\paragraph{YouCook2}
This dataset is a task-oriented, instructional video dataset of 2,000 YouTube videos showing a third-person perspective of individuals cooking 89 separate recipes \cite{zhou2017automatic}. On average, each recipe has 22 videos. The recipe steps for each video are annotated with temporal boundaries and described by imperative English sentences. This dataset is often evaluated in conjunction with MSRVTT as it requires more temporal information in comparison \cite{Bain21,buch2022revisiting,robustness_multimodal}.

\paragraph{YouTube-8M}
This dataset \cite{abuelhaija2016youtube8m} is a collection of 8 million YouTube videos with 4,716 classes. These are categorized into 24 topics: including sports, games, arts, and entertainment.

\subsubsection{Audio-Based Datasets}
\begin{table}
    \centering
    \caption{\textbf{Audio} based datasets that are used for multimodal self-supervised representation learning with video.
    }
    \small
    \resizebox{0.85\textwidth}{!}{\begin{tabular}{cccccc}
        \toprule
        \textbf{Dataset} & \textbf{Labels} & \textbf{Modalities} & \textbf{Classes} & \textbf{Audio Clips} & \textbf{Tasks}  \\ 
        \hline
         AudioSet (AS) \cite{audioset}  & Event & Video+Audio & 632 & 2M & Classification \\
        \hline
        DCASE \cite{dcase}  & Event & Audio & 11 & 20 & Classification \\
        \hline
        ESC-50 \cite{piczak2015esc}  & Activity & Audio& 50 & 2,000 & Classification \\
        \hline
        LRS2 \cite{Chung_2017_CVPR}  & - & Video+Audio & -   & 96,318 &  Action-Recognition \\
        \hline
        VGG-Sound \cite{chen2020vggsound} & Event & Video+Audio & 310 & 210,00 & Classification \\
        \hline
       
        VoxCeleb \cite{voxceleb2}  & - & Video+Audio & -  & 6,000 & Speaker Verification \\
        \bottomrule
    \end{tabular}}
    
    \label{tab:audio_datasets}
\end{table}

While the video datasets used in multimodal problems have associated text, they do not always have associated audio. Additional audio datasets are typically used when the approach includes audio but the video-based dataset does not have audio as a signal.
Table \ref{tab:audio_datasets} lists audio based datasets which are commonly used for multimodal self-supervised representation learning with video. These datasets are used for tasks such as event classification, text-to-video retrieval, text-to-audio retrieval, video-captioning, video description, and text summarization.

\paragraph{AudioSet}
This dataset \cite{audioset} is a collection of over 2 million human-annotated user-generated clips from YouTube. These clips are 10-seconds each and have an ontology of 632 event classes. This dataset is a multi-label dataset, meaning one clip can have more than one 
activity class.

\paragraph{DCASE}
This dataset \cite{dcase} is a small collection of 20 short mono sound files for 11 sound classes with annotations of event temporal boundaries. This dataset is most commonly used for event classification.

\paragraph{ESC-50}
This dataset \cite{piczak2015esc} is a collection of 2000 audio clips that are 5 seconds along. There are a total of 50 classes used for event classification. 

\paragraph{LRS2}
This dataset \cite{Chung_2017_CVPR} is a collection of lip reading from news and talk shows. Each lip reading sentence is 100 characters long with 96,318 utterances in total. 

\paragraph{VGG-Sound} This dataset \cite{chen2020vggsound} is a collection of 200,000 short videos from YouTube where the audio source is visually present and consists of 310 audio events. This dataset is used as a multimodal dataset for joint audio and visual tasks. 

\paragraph{VoxCeleb}
This dataset \cite{voxceleb2} is a collection of over 1 million utterances from 6,112 celebrities extracted from user-generated videos on YouTube. This dataset is also multilingual covering 145 different nationalities. This dataset is most commonly used for visual speech synthesis, speech separation, cross-modal transfer and face recognition.

%% file: sections/pretext.tex
A pretext task is a self-supervised training objective that uses \textbf{pre-defined tasks for the network to solve} in order to learn representations that can be used later for downstream tasks. The idea is, if a model is able to solve a complicated task that requires a high-level understanding of its input, then it will learn more generalizable features. 
Emerging from the image domain, these tasks often relate to appearance statistics including context prediction \cite{doersch2015unsupervised}, colorization \cite{deshpande2015learning, iizuka2016let, larsson2016learning, larsson2017colorization}, ordering shuffled image patches \cite{carlucci2019domain, goyal2019scaling, mundhenk2018improvements, noroozi2016unsupervised, noroozi2018boosting,  Wei_2019_CVPR, kim2018learning}, and geometric transformation recognition \cite{gidaris2018unsupervised}. 
These appearance-based tasks have also been extended to the video domain \cite{jing2019selfsupervised, tao2020selfsupervised, Wang_2019_CVPR_pretext, noroozi2016unsupervised, ahsan2018video, ijcai2021, kim2019selfsupervised}, often coupled with additional tasks that are specific to motion dynamics in videos. Video-specific tasks are related to the time domain, such as playback speed \cite{Yao_2020_CVPR, wangjianglie2020, jenni2020video, Benaim_2020_CVPR, 9412071} and temporal order \cite{Fernando_2017_CVPR, Xu_2019_CVPR, 10.1007/978-3-319-46448-0_32, Lee_2017_ICCV}. Because these tasks require a greater understanding of activities and their relation to time, the learned representations generalize well to other video tasks.
In summary, this survey covers the following pretext tasks: 1) appearance statistics \cite{jing2019selfsupervised, Wang_2019_CVPR_pretext}, 2) temporal order \cite{Fernando_2017_CVPR, Xu_2019_CVPR, 10.1007/978-3-319-46448-0_32, Lee_2017_ICCV, wei2018learning, pathak2017learning}, 3) jigsaw \cite{noroozi2016unsupervised, ahsan2018video, ijcai2021, kim2019selfsupervised}, 4) and playback speed \cite{Yao_2020_CVPR, wangjianglie2020, jenni2020video, Benaim_2020_CVPR, 9412071}.

\subsubsection{Appearance Statistics Prediction} 
\begin{figure*}
    \centering
    \includegraphics[width=.75\linewidth]{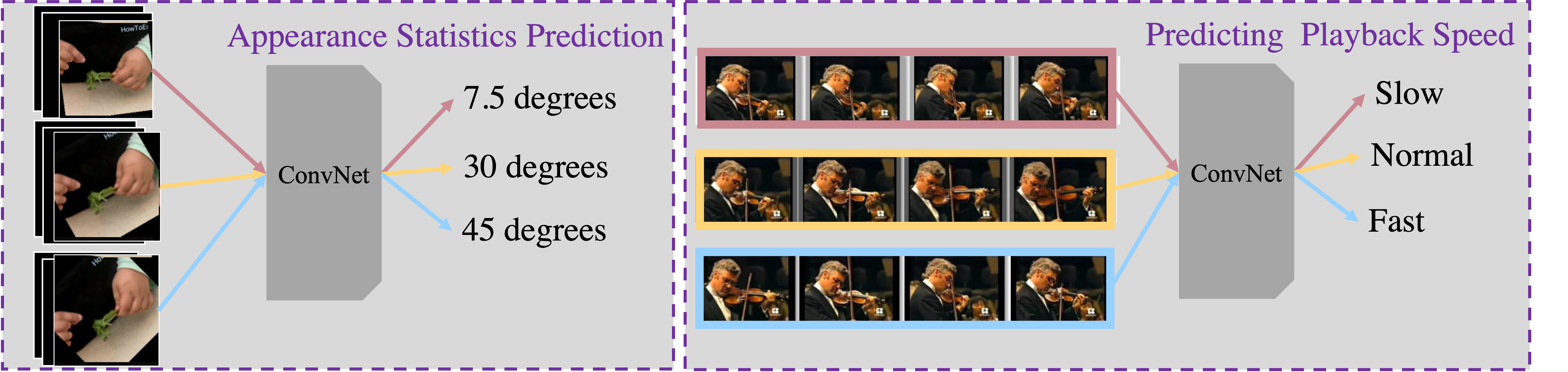}
    \caption{Toy examples of pretext learning tasks. The appearance statistics prediction example (left) shows a series of frames augmented by rotation and the model is asked to predict the degree of rotation used. Similarly, in the playback speed prediction example (right) augmentation is applied on the temporal dimension by changing the speed in which a clip is played and the model is asked to predict the magnitude of that change.}
    \label{fig:pretext_playback}
\end{figure*}

In this task, a model is asked to predict, or classify, an appearance modifying augmentation applied to a video clip. Example augmentations could be color, rotation, and random noise. Rotation based augmentation is inspired from the image domain in \cite{jing2019selfsupervised}. This image based approach was extended to video \cite{jing2019selfsupervised}, where each frame in a video is rotated by four different degrees, separately. The four, different variations are then all passed separately to a 3D CNN. A cross-entropy loss is then used between the predicted rotation and the actual rotation during training. A toy example of this task is shown in Figure \ref{fig:pretext_playback}. While this approach performed well, it only extended the image-based approach to each frame in a video, and did not take into account the temporal dimension. In order to directly address time, \cite{Wang_2019_CVPR_pretext} uses optical flow as a pretext task. Using vertical and horizontal grid lines, the model predicts motion and spatial statistics according to the locations in the frame's grid. For example, the model is asked to predict in a frame where the largest motion statistic is, and where the dominant orientation statistic is. 

\subsubsection{Playback Speed} 
This task classifies the playback speed of an augmented clip \cite{Yao_2020_CVPR, wangjianglie2020, jenni2020video, Benaim_2020_CVPR, 9412071}. This task often starts by taking clips of $t$ frames from each video $V \in \mathbb{R}^{T \times C \times H \times W}$ and selecting frames in a way that the playback speed is altered. This is done by collecting every $p$ frames, where $p$ is the playback rate, either speeding up the video or slowing it down. As illustrated in Figure \ref{fig:pretext_playback}, the task is to \textit{classify} $p$ or to \textit{reconstruct} a clip at the original speed $p$. A simple approach was proposed in \cite{jenni2020video} that adds a motion-type permutation that results in either the modified speed, random, periodic or warped transformations. The task has two multi-class classifications, one detecting which transformation was used and the other detecting speed $p$. To improve the performance, later approaches utilize additional tasks to the classification of $p$. For example, in \cite{Yao_2020_CVPR} the authors introduced a reconstruction loss as an additional signal. The model would encode a modified version of a clip and use the encodings to reconstruct the clip at its original playback rate. However, in \cite{wangjianglie2020} a contrastive loss performed better as an additional signal. This method defined the anchor as the original clip, positive samples are a modified speed of the same clip, and negative samples are clips from other videos. 

\begin{figure}
    \centering
    \includegraphics[width=.75\linewidth]{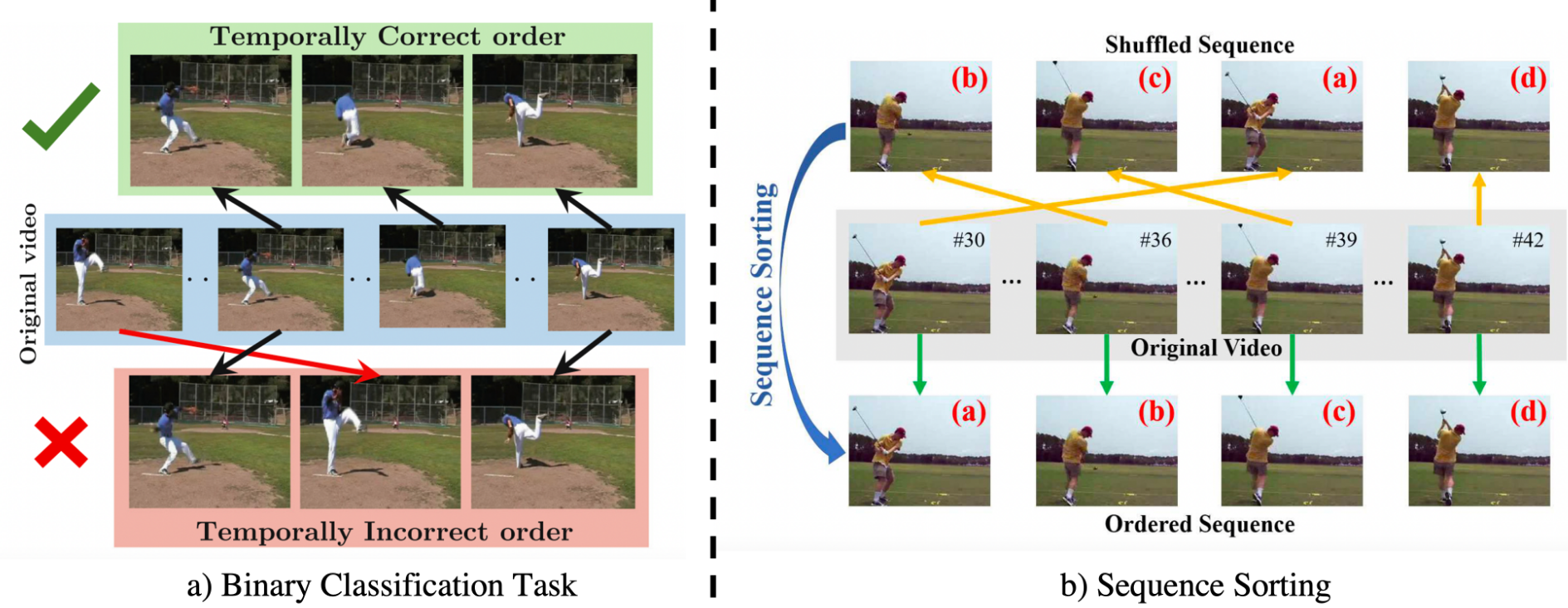}
    \caption{Examples of frame ordering tasks. Images in (a) show order classification from \cite{10.1007/978-3-319-46448-0_32}. This task encodes both the original sequence of frames and a shuffled version and the model must classify a clip on whether it is the correct or shuffled version. Images in (b) show sequence sorting from \cite{Lee_2017_ICCV}. This task takes a shuffled version of a clip and must predict the correct order of the frames.}
    \label{fig:frame_ordering}
\end{figure}

\subsubsection{Temporal Order} 
In temporal order classification, each video $V$ is split into clips of $t$ frames. 
Each set of clips contains a single clip in the correct order, and the remaining clips are modified by shuffling the order. For example, $(t_2, t_1, t_3)$ is incorrect while $(t_1, t_2, t_3)$ is correct \cite{Fernando_2017_CVPR, Xu_2019_CVPR}. The task is called \textit{odd-one-out learning}, defined in \cite{Fernando_2017_CVPR}, where a model is trained to classify whether the input clip is in the correct order, most often using a binary-classifier (see Figure \ref{fig:frame_ordering}a). When using frame ordering, the difference between two frames is not enough to recognize a change in motion for some activities. To address this issue, \cite{Xu_2019_CVPR} extracts short sub-clips and shuffles their order. This clip-based order prediction allows for better comparison because the dynamics of an action are maintained in a sub-clip. To further improve retention of motion dynamics, time windows are extracted where the motion was at its highest, to ensure the model would be able to differentiate between two frames and their motion \cite{10.1007/978-3-319-46448-0_32, Lee_2017_ICCV}. More specifically, \cite{10.1007/978-3-319-46448-0_32} extracts tuples of three frames for each video sampled from varying temporal windows. The frames selected are those that contain the largest amount of motion, computed by optical flow. Negative samples are tuples that are out of order, and positive samples are those in order, including a reverse order e.g. $(t_3, t_2, t_1)$. The model is trained using the \textit{odd-one-out} strategy. The authors in \cite{Lee_2017_ICCV} improved learning further by sorting the frames using pairwise feature extraction on each pair of frames, rather than \textit{odd-one-out}. This approach predicts the actual clip order using shuffled high-motion clips as input (shown in Figure \ref{fig:frame_ordering}b).

\subsubsection{Video Jigsaw} 
\begin{figure}
    \centering
    \includegraphics[width=\linewidth]{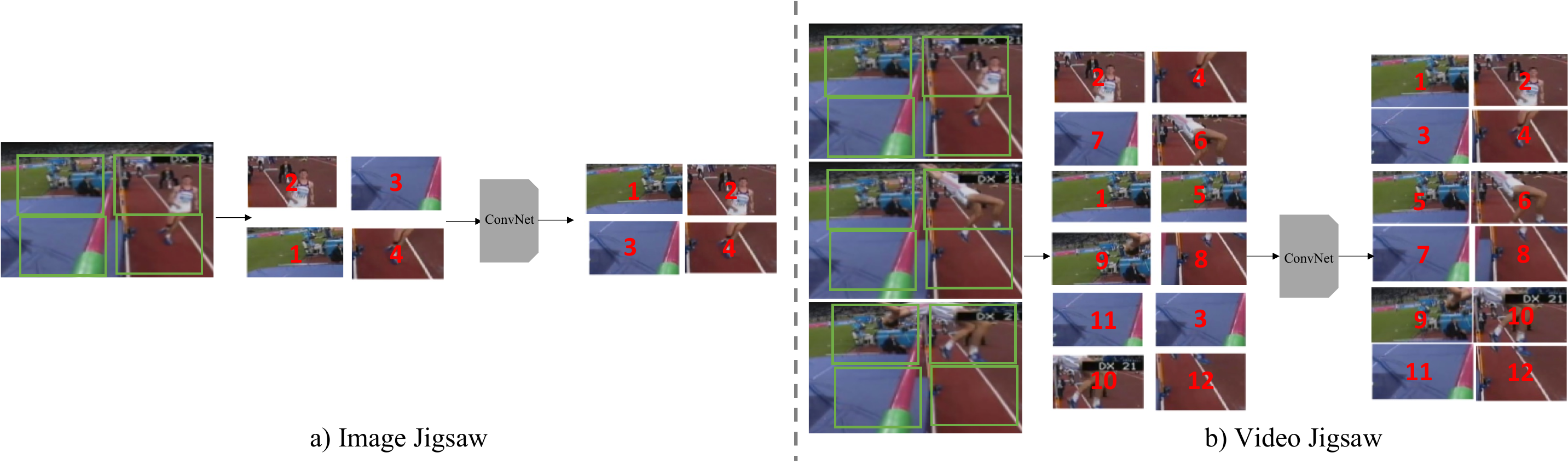}
    \caption{An example of the jigsaw pretext task for images \cite{noroozi2016unsupervised} compared to an example of the jigsaw pretext task for video from \cite{ahsan2018video}. a) An example of image jigsaw using a frame from the UCF101 dataset \cite{soomro2012ucf101}, where the process starts with splitting image(s) into a grid and extracting the pixels from those grids into isolated patches. These patches are then shuffled and the model must predict the correct ordering of the patches. b) An example of video jigsaw using multiple frames, where the process starts by splitting each frame into a 2D grid or splits multiple frames into 3D grids. The patches are then shuffled and the model is asked to predict the correct order or the permutation used to shuffle the patches. }
    \label{fig:jigsaw}
\end{figure}

Jigsaw task was first introduced in the image domain \cite{noroozi2016unsupervised} where an image is split into multiple patches and then shuffled. Each patch is assigned a number and the permutation $P$ that is used to change the order of those numbers (e.g. 9, 4, 6, 8, 3, 2, 5, 1, 7). The seminal work \cite{noroozi2016unsupervised} generated the set of possible permutations to choose from by maximizing the Hamming distance between the original patch order and the permuted order. The final list of possible permutations are the top-k most distant permutations, resulting in a final set of permutations $S$. The objective is a multi-class classification problem, where the classifier predicts which permutation $S_k \in S$ was applied to the image. This approach has been extended to video domain in several works \cite{ahsan2018video,ZHAO2020106534,ijcai2021, kim2018learning,kim2019selfsupervised,Cruz_2017_CVPR} where three dimensional patches are used as illustrated in Figure \ref{fig:jigsaw}. 
One of the challenges in extending Jigsaw to videos, is the increased number of patches. The larger the number of patches, the larger the number of permutations to select from when using the image-based approach. In \cite{ahsan2018video}, a video is split into clips of three frames that are treated as one large image, where all the patches in all the frames are shuffled. The permutation sampling strategy was modified by constraining spatial coherence over time. The patches in a given frame were shuffled before shuffling the frames themselves. Combining the jigsaw task with frame ordering, \cite{ZHAO2020106534} handles the increase in permutations by adopting a multi-stream CNN that receives a shuffled sequence of visual samples as input, where a sample in the sequence goes through a different branch.
To address time, \cite{ijcai2021, kim2019selfsupervised} proposed approaches that use a cubic representation of video clips over time and space. In \cite{kim2019selfsupervised}, a video clip is considered as a cuboid which consists of 3D grid cells. The model learns to classify which permutation was used to modify the clip. In order to improve the permutation sampling strategy, \cite{ijcai2021} invoked a constraint where three spatio-temporal dimensions $(T \times W \times H)$ are shuffled sequentially, rather than simultaneously, to improve continuity of the cuboid. Each jigsaw group of pieces is shuffled rather than each jigsaw piece individually. 

\begin{table}[]
    \centering
    \caption{Downstream evaluation of action recognition on pretext self-supervised learning measured by prediction accuracy. Top scores are in \textbf{bold} and second best are \underline{underlined}. Playback speed related tasks typically perform the best.}
    \resizebox{\textwidth}{!}{\begin{tabular}{llllll}
\toprule
Model &     Subcategory & Visual Backbone &     Pre-Train &                      UCF101 &                      HMDB51 \\
\midrule
Geometry \cite{ gan2018geometry}                  &      Appearance &   AlexNet &           UCF101/HMDB51 &              54.10 &              22.60 \\
Wang et al. \cite{Wang_2019_CVPR_pretext}         &      Appearance &       C3D &                  UCF101 &              61.20 &              33.40 \\
3D RotNet \cite{jing2019selfsupervised}           &      Appearance &   3D R-18 &                     MT \cite{monfort2019moments}  &              62.90 &              33.70 \\
\hline
VideoJigsaw \cite{ahsan2018video}                 &          Jigsaw &  CaffeNet &                Kinetics &              54.70 &              27.00 \\
3D ST-puzzle \cite{kim2019selfsupervised}         &          Jigsaw &       C3D &                Kinetics &              65.80 &              33.70 \\
CSJ \cite{ijcai2021}                              &          Jigsaw &   R(2+3)D &  Kinetics+UCF101+HMDB51 &              79.50 &  \underline{52.60} \\
\hline
PRP \cite{Yao_2020_CVPR}                          &           Speed &       R3D &                Kinetics &              72.10 &              35.00 \\
SpeedNet \cite{Benaim_2020_CVPR}                  &           Speed &     S3D-G &                Kinetics &              81.10 &              48.80 \\
Jenni et al. \cite{jenni2020video}                &           Speed &   R(2+1)D &                  UCF101 &  \underline{87.10} &              49.80 \\
PacePred \cite{wangjianglie2020}                  &           Speed &     S3D-G &                  UCF101 &     \textbf{87.10} &     \textbf{52.60} \\
\hline
Shuffle\&Learn \cite{10.1007/978-3-319-46448-0_32} &  Temporal Order &   AlexNet &                  UCF101 &              50.90 &              19.80 \\
OPN \cite{Lee_2017_ICCV}                          &  Temporal Order &     VGG-M &                  UCF101 &              59.80 &              23.80 \\
O3N \cite{Fernando_2017_CVPR}                     &  Temporal Order &   AlexNet &                  UCF101 &              60.30 &              32.50 \\
ClipOrder \cite{Xu_2019_CVPR}                     &  Temporal Order &       R3D &                  UCF101 &              72.40 &              30.90 \\
\bottomrule
\end{tabular}}
    
    \label{tab:pretext_action_recognition}
\end{table}

\begin{wrapfigure}{R}{0.5\textwidth}
    \includegraphics[width=.49\textwidth]{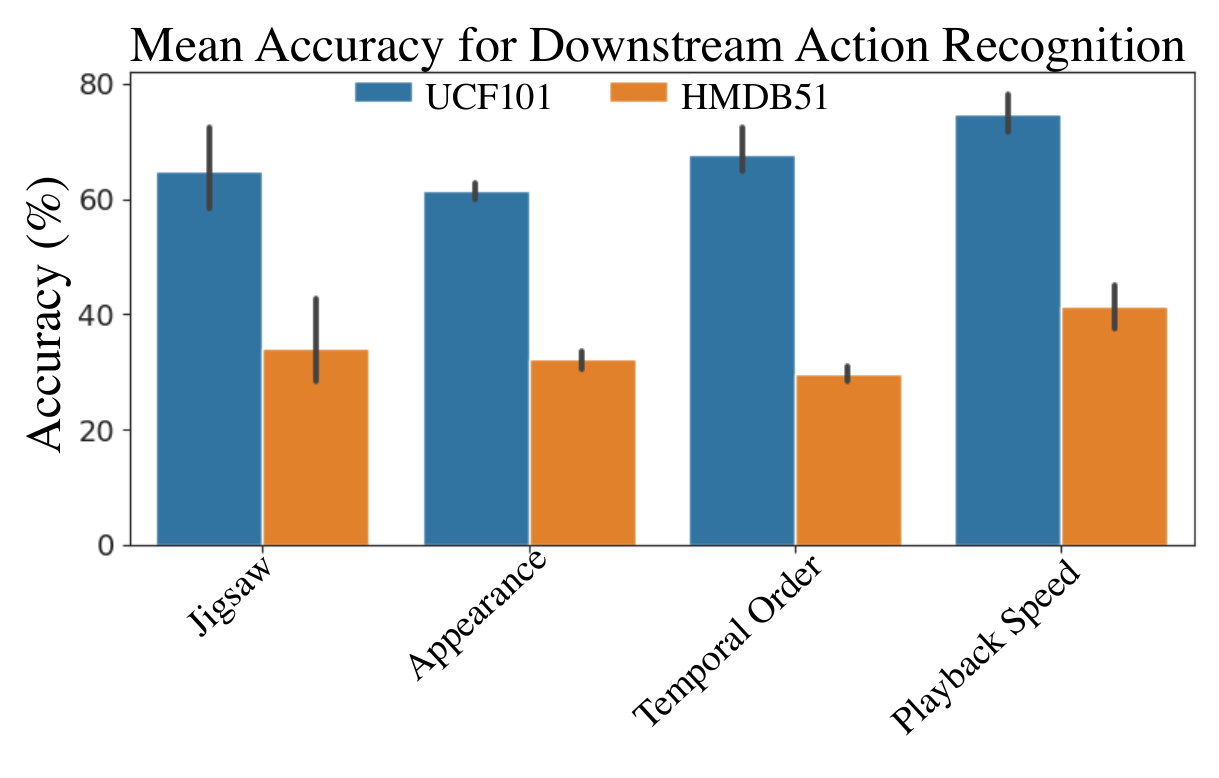}
    \caption{Mean performance on action recognition for video-only pretext learning approaches from Table \ref{tab:pretext_action_recognition}. Using temporal-based approaches, like Temporal Order and Speed, typically outperforms appearance-based approaches.}
    \label{fig:ssl_pretext_video_only}
\end{wrapfigure}

\begin{table}
    \centering
    \caption{Performance for the downstream video retrieval task with top scores for each category in \textbf{bold} and second best scores \underline{underlined}. K/U/H indicates using all three datasets for pre-training, i.e. Kinetics, UCF101, and HMDB51.}
    \resizebox{\textwidth}{!}{\begin{tabular}{llllllll|llll}
\toprule
\multirow{2}{*}{Model} &    \multirow{2}{*}{Category} &        \multirow{2}{*}{Subcategory} & \multirow{2}{*}{Visual Backbone} &               \multirow{2}{*}{Pre-train} &    \multicolumn{3}{c}{UCF101}  &          \multicolumn{3}{c}{HMDB51} \\
 &              &                    &           &                         & R@1                   &       R@5             &     R@10               &        R@1         &          R@5          &        R@10            \\

\hline
SpeedNet \cite{Benaim_2020_CVPR}   &      Pretext &              Speed &     S3D-G &                Kinetics &              13.00 &              28.10 &              37.50 &                 -- &                 -- &                 -- \\
ClipOrder \cite{Xu_2019_CVPR}      &      Pretext &     Temporal Order &       R3D &                  UCF101 &              14.10 &              30.30 &              40.00 &               7.60 &              22.90 &              34.40 \\
OPN \cite{Lee_2017_ICCV}           &      Pretext &     Temporal Order &  CaffeNet &                  UCF101 &              19.90 &              28.70 &              34.00 &                 -- &                 -- &                 -- \\
CSJ \cite{ijcai2021}               &      Pretext &             Jigsaw &   R(2+3)D &  K/U/H &              21.50 &              40.50 &              53.20 &                 -- &                 -- &                 -- \\
PRP \cite{Yao_2020_CVPR}           &      Pretext &              Speed &       R3D &                Kinetics &              23.20 &              38.50 &              46.00 &  \underline{10.50} &  \underline{27.20} &  \underline{40.40} \\

Jenni et al. \cite{jenni2020video} &      Pretext &              Speed &   3D R-18 &                Kinetics &  \underline{26.10} &  \underline{48.50} &  \underline{59.10} &                 -- &                 -- &                 -- \\
PacePred \cite{wangjianglie2020}   &      Pretext &              Speed &   R(2+1)D &                  UCF101 &     \textbf{31.90} &     \textbf{49.70} &     \textbf{59.20} &     \textbf{12.90} &     \textbf{32.20} &     \textbf{45.40} \\
\hline
MemDPC-RGP \cite{Han2020}          &   Generative &         Frame Prediction &   R(2+3)D &                Kinetics &  \underline{20.20} &  \underline{40.40} &  \underline{52.40} &   \underline{7.70} &  \underline{25.70} &     \textbf{65.30} \\
MemDPC-Flow \cite{Han2020}         &   Generative &         Frame Prediction &   R(2+3)D &                Kinetics &     \textbf{40.20} &     \textbf{63.20} &     \textbf{71.90} &     \textbf{15.60} &     \textbf{37.60} &  \underline{52.00} \\

\hline
DSM \cite{Wang2020}                &  Contrastive &    Spatio-Temporal &       I3D &                Kinetics &              17.40 &              35.20 &              45.30 &               8.20 &              25.90 &              38.10 \\
IIC \cite{Tao2020}                 &  Contrastive &    Spatio-Temporal &      R-18 &                  UCF101 &              42.40 &              60.90 &              69.20 &              19.70 &              42.90 &              57.10 \\
SeLaVi \cite{Asano2020}            &  Cross-Modal &     Video+Audio &  R(2+1)D  &                Kinetics &              52.00 &              68.60 &     \textbf{84.50} &              24.80 &  \underline{47.60} &     \textbf{75.50} \\

CoCLR \cite{Han2020_cotraining}    &  Contrastive &  View Augmentation &     S3D-G &                  UCF101 &  \underline{55.90} &  \underline{70.80} &  \underline{76.90} &     \textbf{26.10} &              45.80 &  \underline{57.90} \\
GDT \cite{Patrick2020}             &  Cross-Modal &   Video+Audio &  R(2+1)D  &                Kinetics &     \textbf{62.80} &     \textbf{79.00} &                -- &     \textbf{26.10} &     \textbf{51.70} &                -- \\
\bottomrule
\end{tabular}}
    
    \label{tab:video_retrieval_all}
\end{table}

\subsubsection{Discussion}
In pretext self-supervised pre-training, we discussed several techniques: 1) data augmentation on spatial and temporal elements of a video in which the model was asked to make predictions about the related statistics; 2) altering the playback speed of a video in which the model is asked to predict the speed or reconstruct the original speed; 
3) shuffled frames where tasks included predicting the odd-one-out and/or sorting the frames to the original order; 4) shuffled frame patches using a permutation selected from a set of derived, possible permutations in which the model is asked to predict the permutation that was selected. 

Figure \ref{fig:ssl_pretext_video_only} shows the mean accuracy on downstream action recognition for models using pretext learning on both UCF101 \cite{soomro2012ucf101} and HMDB51 \cite{kuehne2011hmdb} datasets. This figure shows that the \textit{pretext tasks relating to playback speed, or temporal consistency, typically perform better}. This is further supported in Table \ref{tab:video_retrieval_all} which shows results for video-retrieval where we observe that playback speed is again outperforming other pretext tasks.
Limitations to these findings is the use of a variety of backbone architectures and pre-training datasets making direct comparison of the proposed approaches difficult.

%% file: sections/generative.tex
\begin{figure}
    \centering
    \includegraphics[width=.95\linewidth]{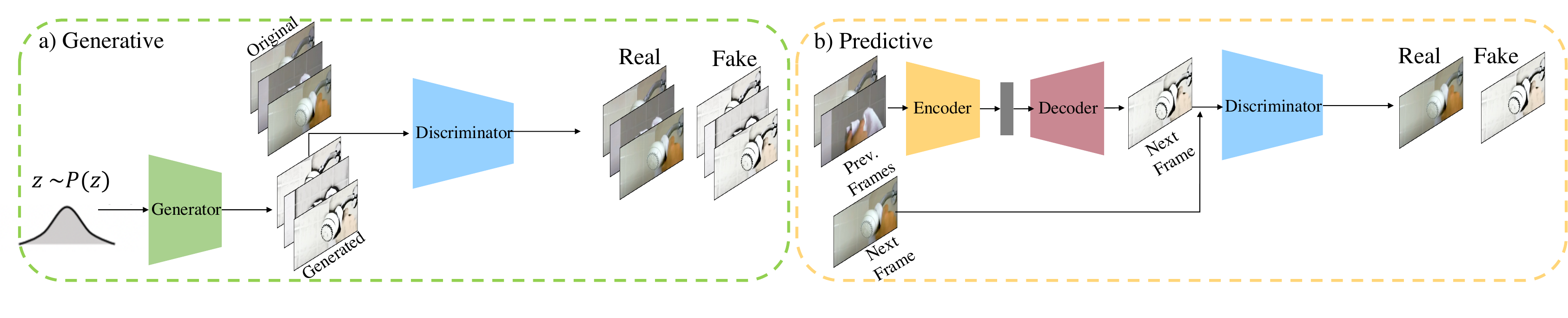}
    \caption{A toy example of generative approaches and their respective loss functions. (a) Shows a generator taking random noise from a normal distribution to generate images. The generated images, and original images, are then passed to a discriminator, which classifies which is real (original) or fake (generated).
    (b) Shows frame predicting where the previous frames are fed to an encoder-decoder to generate the next frame in the sequence. The original frames, and the next frame predictions, are then passed to a discriminator, which classifies which are real (original) or fake (generated).
    }
    \label{fig:generative_works}
\end{figure}

Generative learning utilizes the \textbf{generative power of neural networks to synthesize video or other signals} \cite{vondrick2016generating}. These tasks originated in the image domain, where the most common techniques are Generative Adversarial Networks (GANs) \cite{goodfellow2014generative, zhu2017unpaired} and Masked Autoencoders (MAE) \cite{he2022masked}. These generative approaches for images have been extended to video in \cite{srivastava2015unsupervised, Tulyakov_2018_CVPR, vondrick2016generating,Wei_2022_CVPR,motionmae,feichtenhofer2022masked,wang2022bevt}. \textbf{GANs} involve data synthesis where the synthesized data point is approximated with the actual data point. These are used with an \textbf{encoder-decoder architecture} \cite{goodfellow2014generative}, where an encoder first takes input and projects it to a latent space. A decoder then takes the latent features and projects them to the original space, attempting to reconstruct an original data point. A discriminator is then used to distinguish the generated output from the original data point (see Figure \ref{fig:generative_works}). There can be many variations in the learning objective depending upon the target data point. For example, the target data point could be the input itself \cite{vondrick2016generating} or maybe some unseen point such as future frames in videos \cite{Han2020, Tulyakov_2018_CVPR, Tian2020}. \textbf{Masked Modeling} (MM) is another generative learning framework and is typically used with transformer architectures. 
Similar to masked language modeling (MLM) from the natural language processing (NLP) domain \cite{devlin2018bert, vaswani2017attention}, \textbf{masked modeling} takes a sequence as input, where some input tokens are randomly masked, and the model generates these masked tokens using the context of the other tokens in the sequence (see Figure \ref{fig:mlm}).
For just video, these approaches are MAEs \cite{Wei_2022_CVPR,motionmae,feichtenhofer2022masked,wang2022bevt}. 
With the success of MAEs using only RGB frames and motion, additional signals like audio and text have been shown to further improve generalizability of learned features \cite{Sun2019b, Lei2021, Owens2018, Zhu2020, Xu2020, Li2020, Luo2020}. These approaches are called Multimodal Masked Modeling. 

\subsubsection{Adversarial and Reconstruction}
Typically, GANs are used for video-to-video synthesis tasks and are not often used for self-supervised pre-training. However, \cite{vondrick2016generating} proposed VideoGAN, a generator that uses a two-stream network of 2D and 3D visual features and a discriminator that classifies realistic scenes from synthetically generated scenes and recognizes realistic motion between frames. The methods in \cite{Chen2019,Bansal2018} added time and motion to successful works in image domain \cite{Zhu_2017_ICCV} to extend them to video, although more focused on video-to-video translation. In \cite{Bansal2018}, the cycle-consistency loss, a type of reconstruction loss, from \cite{Zhu_2017_ICCV} was extended to a recycle loss. The original cycle-consistency loss learned two mappings, $G: X \rightarrow Y$ and $F: Y \rightarrow X$ to reduce the space of possible mapping functions by enforcing forward and backward consistency. The \textbf{recycle loss} \textit{utilizes a recurrent loss which predicts future samples, in a temporal stream, given its past, making use of cycle-consistency}. This was further extended in \cite{Chen2019} by focusing on motion. 
This method explicitly employs motion across frames by enforcing motion consistency through optical flow.
The most recent work \cite{Hadji_2021_CVPR} utilized dynamic time warping (DTW) to admit a global cycle consistency loss that verifies correspondences over time. 

\subsubsection{Frame Prediction}
Rather than focusing on reconstructing motion, \cite{Liang2017} and \cite{Tian2020} focused on generating motion from RGB frames and vice versa for future unseen frames. A discriminator and variational autoencoder (VAE) is used to measure the quality of these generated predictions when compared to the ground truth RGB frames and optical flow.
Instead of using optical flow as the motion signal, \cite{Tian2020} proposed the use of encoded features between two frames to decode and extract low-resolution motion maps. High-resolution motion maps are generated using contextual features extracted from spatial regions. Both of these motion maps are then used to make a prediction of the next frame at various resolutions where a \textit{reconstruction loss} is used to measure the quality of the reconstructions. 

To address time directly, \cite{Tulyakov_2018_CVPR} formulates a sequence of frame representations as paths, therefore modeling videos of varying lengths. A recurrent neural network (RNN) constrains the learning of a path in the motion subspace to physically plausible motion. Recurrent structures have also been utilized in \cite{Han2020} which propose Memory-augmented Dense Predictive Coding (MemDPC). MemDPC splits a video into blocks, containing an equal number of frames, and passes them to an encoder to generate embeddings. These embeddings are then aggregated over time using RNNs. During training, a \textit{predictive addressing mechanism} is used to access a memory bank shared between the entire dataset to draw a hypothesis and predict future blocks. 
Another approach \cite{Pan2021} uses a \textbf{temporal mask} to indicate the importance of each frame rather than generating frames. Both the masked set of frames and the original set of frames are sent to a discriminator where the latent representations became similar over training. A \textit{contrastive loss} is used to train the discriminator, measuring the similarity between the query and a set of other samples, called keys, stored in a dictionary \cite{he2020momentum}. 

\subsubsection{Masked Autoencoders}
This approach was originally proposed in the image-domain \cite{he2022masked} where patches of an image are masked. The model encodes the visible patches and decodes both the visible and masked patches. This has been extended to video in several works \cite{Wei_2022_CVPR,motionmae,feichtenhofer2022masked,wang2022bevt} with an emphasis on temporal elements. These approaches typically use a visual-transformer backbone in order to allow for the masking task. MAE \cite{feichtenhofer2022masked} extends image-based Masked Autoencoders to video using spatio-temporal learning. They randomly mask out space-time patches in videos and learn an autoencoder to reconstruct them in pixels. BEVT \cite{wang2022bevt} both extends image based MAE and uses images by utilizing both an image and video stream during pre-training. MotionMAE \cite{motionmae} masks patches and adds emphasis to temporal elements by feeding the encoder output to a Time head and a Space head. These heads process visible and masked tokens for reconstruction of frame patches and motion. Similarly, in MaskFeat \cite{Wei_2022_CVPR} a few space-time cubes of video and the model is asked to predict those using what is remaining. The idea is the model must recognize objects using context as well as how they move. Because these approaches emphasize temporal elements, they have been evaluated on the Something-Something dataset \cite{goyal2017something} as shown in Table \ref{tab:generative_ss_kinetics}.

\begin{table}[]
    \centering
    \caption{Downstream evaluation of action recognition on  self-supervised learning measured by prediction accuracy for Something-Something (SS) and Kinetics400 (Kinetics). SS is a more temporally relevant dataset and therefore is more challenging.  Top scores for each category are in \textbf{bold} and second best scores \underline{underlined}.}
    \resizebox{\textwidth}{!}{\begin{tabular}{lllllll}
\toprule
Model &  Category &   Subcategory & Visual Backbone &     Pre-Train &                      SS &  Kinetics \\ 
\midrule
pSwaV \cite{Feichtenhofer_2021_CVPR} & Contrastive & View Aug. & R-50 & Kinetics & 51.7 & 62.7 \\ 
pSimCLR \cite{Feichtenhofer_2021_CVPR} & Contrastive & View Aug. & R-50 & Kinetics & 52.0 & 62.0 \\
pMoCo \cite{Feichtenhofer_2021_CVPR} & Contrastive & View Aug. & R-50 & Kinetics & 54.4 & 69.0 \\ 
pBYOL \cite{Feichtenhofer_2021_CVPR} & Contrastive & View Aug. & R-50 & Kinetics & 55.8 & 71.5  \\
\hline
BEVT \cite{wang2022bevt} & Generative & MAE & SWIN-B \cite{liu2021swin} & Kinetics+ImageNet \cite{krizhevsky2012imagenet} & 71.4 & 81.1 \\ 
MAE \cite{feichtenhofer2022masked} & Generative & MAE & ViT-H \cite{dosovitskiy2020image} & Kinetics & 74.1 & 81.1 \\ 
MaskFeat \cite{Wei_2022_CVPR} & Generative & MAE & MViT \cite{fan2021multiscale} & Kinetics & 74.4 & \textbf{86.7} \\ 
VideoMAE \cite{tong2022videomae} & Generative & MAE & ViT-L \cite{dosovitskiy2020image} & ImageNet &  \underline{75.3} & \underline{85.1} \\ 
MotionMAE \cite{motionmae} & Generative & MAE & ViT-B \cite{dosovitskiy2020image} & Kinetics & \textbf{75.5} & 81.7 \\

\bottomrule
\end{tabular}}
    
    \label{tab:generative_ss_kinetics}
\end{table}

\subsubsection{Multimodal Masked Modeling}
Masked modeling was first introduced in the natural language processing (NLP) community as Masked Language Modeling (MLM) with the popular Bidirectional Encoder Representations from Transformers (BERT) \cite{devlin2018bert} model. Several works have adapted this approach into multimodal approaches for self-supervised learning with video, typically including additional modalities. A number of these works have used combinations of visual, text and/or audio tokens for masked modeling to learn a joint space of the different modalities. This was first introduced in \cite{Sun2019b} where BERT was extended to video domain by transforming the raw visual data into a discrete sequence of tokens using hierarchical k-means. This allows a combination of textual and visual tokens in a sequence used in a BERT architecture. ASR is used as the text signal in order to add tokens that specify the joining of two sentences {\fontfamily{qcr}\selectfont[SEP]} and the beginning of each sequence {\fontfamily{qcr}\selectfont[CLS]}. For both video and text tokens, the model predicts whether the linguistic sentence is temporally aligned to the visual sentence using the hidden state of the {\fontfamily{qcr}\selectfont[CLS]} token. The temporal alignment task has also been used between video and audio in \cite{Owens2018}. Here the authors proposed a binary classification task between video and audio that is synchronized from the same video, or video with audio that was shifted several seconds.

To improve the cross-modal feature embedding, some recent works proposed a \textbf{masked modeling (MM)} \textit{approach between modalities where tokens from one modality are used to recover tokens from another modality} \cite{Sun2019b, Zhu2020, Xu2020, Li2020, Luo2020}. These approaches utilized multiple pre-training tasks that alter the masked modeling approach and/or combine it with other tasks for learning. Some works even extend modalities into further subcategories such as object and action features: object features are spatial regions in a frame, text object descriptions and action features are verbs in text, and 3D CNN features over a set of frames \cite{Zhu2020}. MM would then be used to predict the tokens from one of the subcategories in one modality using the tokens from the same subcategory in a different modality. 

While these approaches used \textbf{cross-modal learning}, it did not use any \textbf{joint-modal learning}. In order to additionally use joint-learning, the authors in \cite{Luo2020} used both known tokens and other modality tokens to predict masked features in several MM pre-training tasks. To better address the dimension of time in learning, \cite{Li2020} focused on both \textit{local} and \textit{global} levels of video representation where \textit{global} is specific to time. Inspired by \cite{lu2019vilbert, chen2020uniter, vaswani2017attention}, \textit{local} contextual cues were captured using a Cross-Modal Transformer between the subtitles and associated frames. The encodings from this transformer for the entire video clip, rather than frame-wise, were then fed into a Temporal Transformer to learn \textit{global} contextual cues and the final representation. Two of the pre-training tasks were a masked modeling approach, one masking word tokens and the other masking frames.

\label{sec:masked_modeling}

\begin{figure}
    \centering
    \includegraphics[width=.65\linewidth]{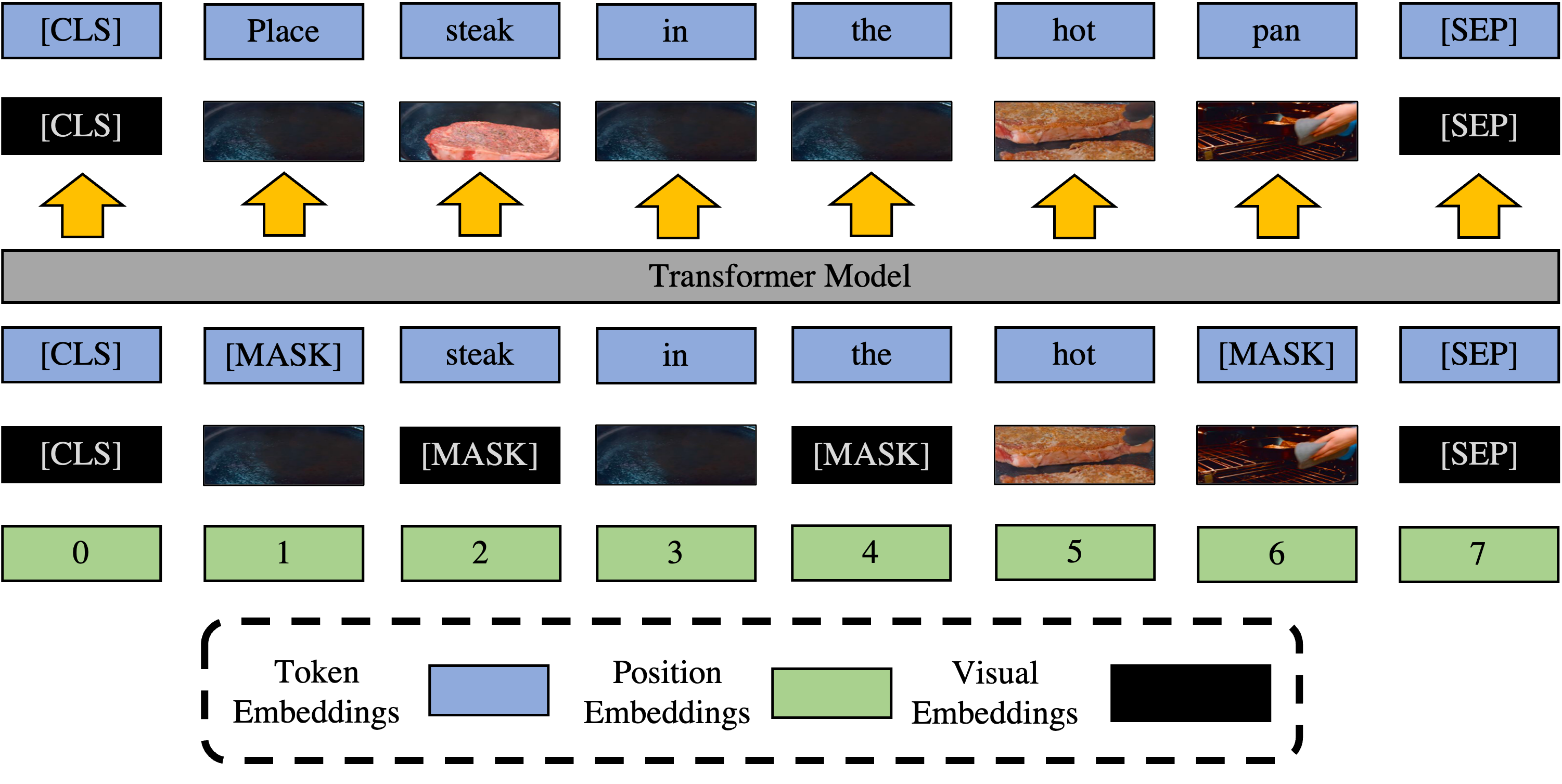}    
    \caption{Inspired by the Masked Language Modelling (MLM) task in NLP, approaches that use masked modeling with multiple modalities convert visual data into discrete tokens that allow a combination of textual and visual tokens in a sequence. In this example, random text and visual tokens are masked and the model is asked to ``fill in the blanks''.}
    \label{fig:videobert}
\end{figure}

\subsubsection{Discussion}

In self-supervised generative learning, we discussed several techniques: 1) the reconstruction of altered input/and or noise; 2) using previous frames to generate future frames in a sequence; 3) generating masked input that can be patches in frames or entire frames; 4) generating masked input of video, text and/or audio.


\begin{table}
\caption{Downstream action recognition evaluation for models that use a generative self-supervised pre-training approach. Top scores are in \textbf{bold} and second best are \underline{underlined}.} 
\label{tab:action_recognition_generative}
    \centering
    \resizebox{.85\textwidth}{!}{
\begin{tabular}{llllll}
\toprule
{} &             Subcategory & Visual Backbone & Pre-train &             UCF101 &             HMDB51 \\
Model                                  &                         &           &           &                    &                    \\
\midrule
Mathieu et al. \cite{mathieu2016deep} &  Frame Prediction &       C3D &  Sports1M &              52.10 &                -- \\

VideoGan \cite{vondrick2016generating} &  Reconstruction &       VAE &   Flickr  &              52.90 &                -- \\

 Liang et al. \cite{Liang2017}         &  Frame Prediction&      LSTM &    UCF101 &              55.10 &                -- \\
VideoMoCo \cite{Pan2021}               &  Frame Prediction &   R(2+1)D &  Kinetics &              78.70 &              49.20 \\
MemDPC-Dual \cite{Han2020}             &  Frame Prediction &   R(2+3)D &  Kinetics &  \underline{86.10} &  54.50 \\
Tian et al. \cite{Tian2020}            &  Reconstruction &  3D R-101 &  Kinetics &     88.10 &     \textbf{59.00} \\
VideoMAE \cite{tong2022videomae}  & MAE & ViT-L \cite{dosovitskiy2020image} & ImageNet &  \underline{91.3} & 62.6 \\
MotionMAE \cite{motionmae}  & MAE & ViT-B \cite{dosovitskiy2020image} & Kinetics &  \textbf{96.3} & -- \\

\bottomrule
\end{tabular}}\end{table}

Generative approaches evaluated on action-recognition for the UCF101 and HMDB51 dataset are shown in Table \ref{tab:action_recognition_generative}. We observe that the approaches that use transformers to generate masked input as their pre-training task are the highest performers. Of the approaches that use the frame prediction and/or reconstruction tasks, methods that use temporal elements like motion and frame sequence are higher performers \cite{Tian2020,Han2020}.

Generative approaches evaluated on action-recognition with the more temporal specific Something-Something (SS) \cite{goyal2017something} and Kinetics400 (Kinetics) dataset are shown in Table \ref{tab:generative_ss_kinetics}. The focus on using visual-transformers and temporal-specific learning mechanisms shows a large performance increase for both SS and Kinetics. 

Because Multimodal masked modeling (MM) uses text, these generative approaches can additionally be evaluated on caption generation. Generative approaches evaluated on caption generation are shown in Table \ref{tab:video_caption} on the YouCook2 dataset \cite{zhou2017automatic}. We observe better performance when the models are pre-trained on the HowTo100M dataset. Additionally, MM shows high performance without pre-training/fine-tuning on the target dataset.



\begin{table}[h]
\caption{Downstream evaluation for video captioning on the YouCook2 dataset for video-language models. Top scores are in \textbf{bold} and second best scores are \underline{underlined}.
MM: Masked modeling with video and text, and K/H: Kinetics+HowTo100M.
}
\label{tab:video_caption}
\resizebox{0.99\textwidth}{!}{\begin{tabular}{lllllllllll}
\toprule
 &      &       & \multicolumn{2}{c}{Backbone} &      &      &     &      &    \\
Model &     Category &      Subcategory &Visual & Text &           Pre-train &     BLEU3 &     BLEU4 &    METEOR &     ROUGE &    CIDEr \\

\midrule
CBT \cite{Sun2019a}       &  Cross-Modal &       Video+Text &     S3D-G &    BERT &        Kinetics &                -- &   \underline{5.12} &  \underline{12.97} &  \underline{30.44} &     \textbf{0.64} \\
COOT \cite{COOT}          &  Cross-Modal &       Video+Text &     S3D-g &     BERT &        YouCook2 &     \textbf{17.97} &     \textbf{11.30} &     \textbf{19.85} &     \textbf{37.94} &               -- \\
\hline

VideoBert \cite{Sun2019b} &      Generative &  MM &     S3D-g &   BERT &          Kinetics &               7.59 &               4.33 &              11.94 &              28.80 &              0.55 \\
ActBERT \cite{Zhu2020}    &      Generative &  MM &   3D R-32 &  BERT & K/H &               8.66 &               5.41 &              13.30 &              30.56 &              0.65 \\

VLM \cite{Xu2020}         &      Generative &  MM &     S3D-g &    BERT &        How2 &  \underline{17.78} &  \underline{12.27} &  \underline{18.22} &  \underline{41.51} &  \underline{1.39} \\
UniVL \cite{Luo2020}      &      Generative &  MM &     S3D-g &     BERT &       How2 &     \textbf{23.87} &     \textbf{17.35} &     \textbf{22.35} &     \textbf{46.52} &     \textbf{1.81} \\
\bottomrule
\end{tabular}}
\end{table}

%% file: sections/contrastive.tex
\begin{figure}
    \centering
    \includegraphics[width=.85\linewidth]{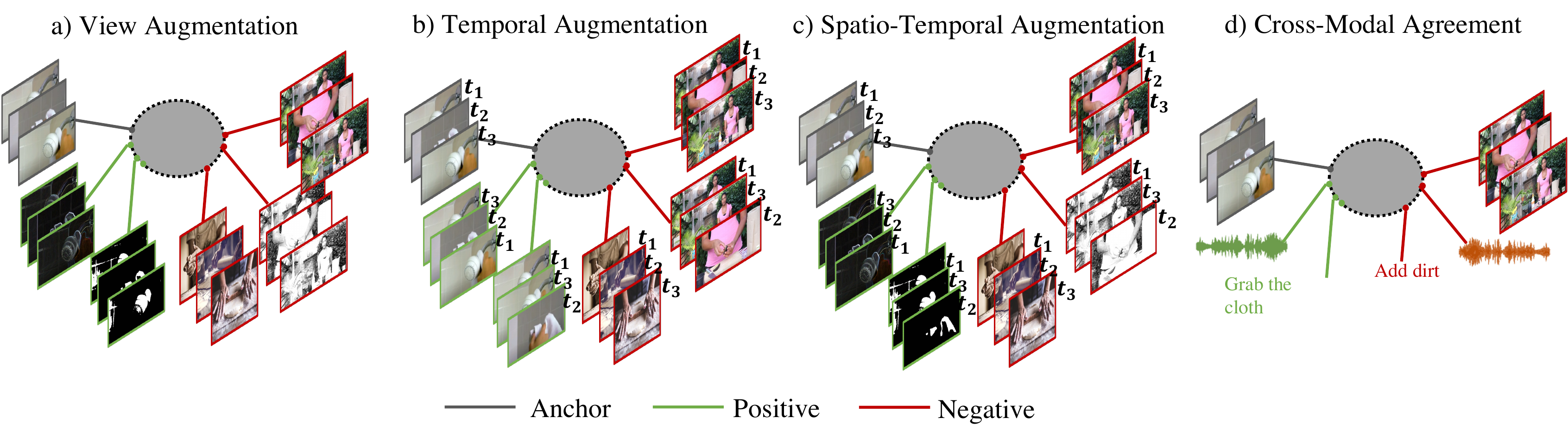}
    \caption{Toy examples of how positive samples are generated for a contrastive loss aiming to bring an anchor sample (grey) closer to the positive samples (green) in latent space while pushing negative samples (red) further away. In a) view augmentation, positive examples are perturbed versions of the anchor clip using color changes and/or geometric changes for each frame. In b) temporal augmentation positive examples are generated by shuffling the order of frames c) and spatio-temporal is a combination of view and temporal changes. The last example d) shows cross-modal agreement where positive samples are audio or text samples from the anchor clip and negative samples are audio, video and/or text from other clips.}
    \label{fig:contrative_learning_examples}
\end{figure}

Contrastive learning provides \textit{a self-supervised way to push positive input pairs closer together while pushing negative input pairs farther apart}.  Figure \ref{fig:contrative_learning_examples} shows how a positive pair is generated from an anchor frame, in a video, and a frame at a different time in the same video (e.g. an adjacent frame in the same clip). To generate a set of negative samples, pairs are created between the anchor frame and frames from other videos. There are numerous ways of generating positive and negative samples which is the main differentiating factor between contrastive approaches. 

There are three main objectives to contrastive learning: binary cross-entropy \cite{Knights2020, Yao2020}, discriminator \cite{Tian2020, Wang2020, Tao2020, Xue2020}, and Noise-Contrastive Estimation (NCE) \cite{mnih2013learning, Qian2020, Tian2020, Yao2020, Han2020_cotraining, Tian2020, Knights2020, Hjelm2020, Lorre2020}. \textbf{Cross-entropy} is used with a binary classifier to determine whether a pair is ``good" or not. \textbf{Discriminators} are used to return scores that are higher for positive pairs and smaller for negative pairs. The most commonly used contrastive training objectives are variations of the NCE and originated from the NLP domain. The \textbf{NCE loss} \textit{takes as input a positive pair of samples and a set of negative samples, minimizing the distance between the positive pair and maximizing the distance between the negative pairs} \cite{nce, smith2005contrastive, nce_manyneg, collobert2008unified, oord2018representation}.

The majority of these works vary how positive and negative pairs are generated to minimize and maximize the distance between them respectively.  In the image domain, this is \textit{traditionally done by augmenting an image in different ways to generate positive samples} \cite{wu2018unsupervised, ye2019unsupervised, he2020momentum, chen2020simple, misra2020self}. Such augmentations include  rotation, cropping, random grey scale and color jittering \cite{ye2019unsupervised, chen2020simple}. To extend these works in video can be difficult because each videos comparison adds to the memory required, especially if using multiple augmentations for multiple positive samples. Another challenge is including time in the augmentations. Some approaches simply apply the same augmentations as in images to each frame \cite{Hjelm2020, Han2020_cotraining, Tian2020_CMC, Feichtenhofer_2021_CVPR}. Some approaches include additional permutations that may be temporal based, such as shuffling frames \cite{Knights2020,Lorre2020}. Finally, some rely on motion and flow maps as the positive samples \cite{Rai_2021_CVPR}. 

Approaches also vary in how they maintain collections of negative pairs in order to keep computation and space low, typically using a \textit{memory bank} or \textit{momentum encoder} \cite{technologies9010002}. To further improve the memory use and performance, contrastive learning expanded into using other modalities from video that includes audio and/or text \cite{patrick2020support, Amrani2020, xu-etal-2021-videoclip, Miech2019, Miech2020}. Approaches that are referred to as multimodal or cross-modal learning are referring to learning between video, audio and/or text. Text is a common secondary modality because the NCE loss originated in the NLP domain \cite{mnih2013learning} with great success. Text also adds semantic knowledge something that would otherwise require in-depth annotations for video. To train using these signals, the existing approaches often use pair-wise comparisons between separate embeddings of each signal.

The Contrastive approaches used for self-supervised learning in the video domain are split into four main categories: \textit{view augmentation}, \textit{temporal augmentation}, \textit{spatio-temporal augmentation}, \textit{multimodal alignment} \textit{clustering}, \textit{multimodal clustering} and \textit{multimodal cross-modal agreement}.


\subsubsection{View Augmentation} 

Originally proposed in the image-domain, view augmentation, or data augmentation, \textit{focuses on changes in appearance through various transformations}. These include random resized crop, channel drop, random color jitter, random grey, and/or random rotation \cite{wu2018unsupervised, ye2019unsupervised, he2020momentum, chen2020simple, misra2020self}. In video, these augmentations are applied to each frame in a clip \cite{Hjelm2020, Tian2020_CMC, Han2020_cotraining}. Positive pairs are augmented versions of the original clip while negative pairs are clips from other videos. Each type of augmentation results in a different ``view" of the original clip. 

To extend to videos, \cite{Feichtenhofer_2021_CVPR} applied the popular image-based approaches MoCo \cite{he2020momentum}, SimCLR \cite{chen2020simple}, BYOL \cite{grill2020bootstrap} and SwAV  \cite{caron2021unsupervised} by using the same perturbations across a series of frames. They showed that the approaches performed well when a temporal persistency objective was added. More specific to video, and utilizing an encoder-decoder network, \cite{Hjelm2020} combined augmentations with a view generator that split clips into two down-sampled sequences. These sequences are then passed to a decoder that processes global features at later layers and local patches at earlier layers. An NCE loss is applied to the individual layers across all locations of the encoder, maximizing the mutual information across all locations over all layers. Instead of relying only on an NCE loss, \cite{Tian2020_CMC} additionally uses a discriminator to predict high values for positive pairs and low values for negative pairs. 

\subsubsection{Temporal Augmentation} 
A more video-specific augmentation is temporal. It is used to generate pairs from modifying the temporal order or the start and end point of a clip interval \cite{Knights2020,Lorre2020}. In \cite{Knights2020}, temporal coherency is used by maximizing a similarity function between two temporally adjacent frames in the same video, while minimizing similarity between frames from other videos. The NCE loss uses multiple negative samples that are selected by their embedding distance outside a hypersphere. As training progresses, the number of selected negative samples is increased, which results in a random sampling. 
Rather than using the next single frame as a positive, \cite{Lorre2020} uses an autoregressive model to generate a sequence of future frames as positive samples. 
The model is then trained using an NCE objective which uses the actual latent representations as an anchor, the generated representations as positive samples, and the generated latent representations from other videos as negative.

\subsubsection{Spatio-Temporal Augmentation}
While temporal augmentations perform well, some recent works which incorporate both time and space have been found to be more effective. In \cite{Han2020_cotraining} an encoder for augmented RGB frames and an encoder for augmented optical flow are used. The encoders are trained using a modified InfoNCE loss that compares a given sample to multiple positive samples by finding the top-k most similar samples. It first mines for samples of the same type, e.g. RGB to RGB, and then switches them, e.g. RGB samples are paired with the top-k most similar flow samples and vice versa.
The authors in \cite{Yao2020} propose a two pairing strategy: inter-frame instance discrimination using NCE, which highlights temporal elements, and intra-frame instance discrimination using cross-entropy which highlights spatial. 
Similarly in \cite{Wang2020}, compares multiple signals using InfoNCE, where temporal and spatial are considered separately and jointly. Three versions of each clip are generated from 1) basic augmentation 2) Thin-Plate-Spline (TPS) \cite{jaderberg2015spatial, shi2016robust} generated spatial local disturbance and 3) pace changing with optical-flow scaling combined with temporal shift for temporal local disturbance (TLD). 
In order to focus more on the spatio-temporal relationship rather than the two separately, \cite{Qian2020} uses both temporal and spatial augmentation to generate positive pairs for an InfoNCE loss \cite{chen2020simple}. Inspired by this, \cite{Tao2020} used multi-view augmentation to generate positive samples while negative samples were generated from frame repeating or frame shuffling to distort the original temporal order. 
To focus on both local and global representations, \cite{DAVE2022103406} uses a local loss that treats non-overlapping clips as negatives and spatially augmented versions of the same clip as positives.


\begin{figure}
    \centering
    \includegraphics[width=.85\linewidth]{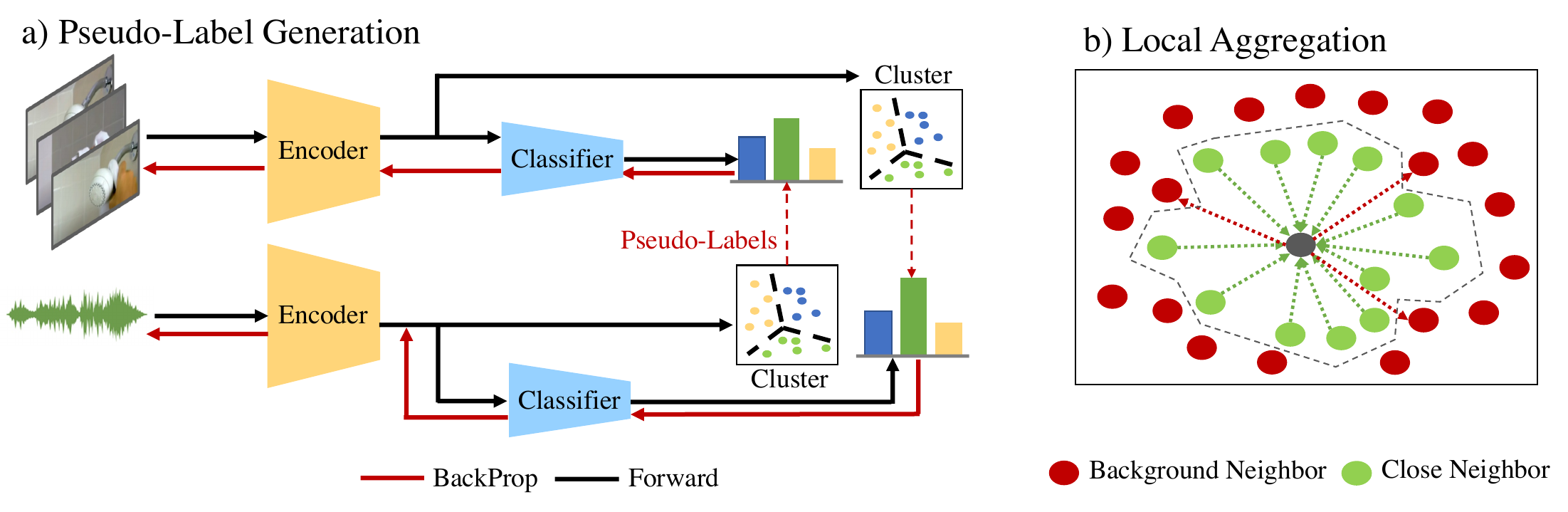}
    \caption{Toy examples of how clustering is used in self-supervised learning for video. a) Pseudo-Label generation uses cluster assignments as pseudo labels with multiple modalities as in \cite{Alwassel2020}. The cluster assignments from one signal are used as pseudo-labels for the other. b) Local Aggregation focuses on the latent space where for each sample ``close'' neighbors are pulled closer and ``background'' neighbors are pushed further away like in \cite{zhuang2019local,he2017mask,wang2013action,tokmakov2020unsupervised}.
    }
    \label{fig:xdc}
\end{figure}

\subsubsection{Clustering}
Clustering is an approach to self-supervised learning that \textit{focuses on the optimal grouping of videos in a cluster}. In the image-based approach \cite{zhuang2019local} two sets of neighbors are defined for each sample: background neighbors $\mathbf{B}_i$, and close neighbors $\mathbf{C}_i$.  The model learns to make $\mathbf{C}_i$ closer to a video's representation, while $\mathbf{B}_i$ is used to set the distance scale that defines ``closeness". In \cite{Zhuang_2020_CVPR}, this approach was extended to videos using a two-branch architecture: one that embeds static, single frames and another that embeds dynamically, multiple frames. 
At each step, all videos are compared and processed via loss functions adapted from \cite{zhuang2019local, wu2018unsupervised}: Instance Recognition (IR) and Local Aggregation (LA). The LA loss function use dynamic online k-means clustering to find ``close” neighbors. This two branch approach is also utilized in \cite{tokmakov2020unsupervised} where the first branch use a 3D ConvNet to embed video features, while the second branch embed the trajectory of videos using dimensionality reduction on IDT descriptors \cite{wang2013action} of Mask-RCNN human detections \cite{he2017mask}. Both branches use k-means clustering to identify close neighbors and background neighbors as described in \cite{zhuang2019local}. After several epochs of training, joint IDT and 3DConvNet embeddings are generated by concatenating their respective features for the k-means clustering objective.
To improve clustering further, the authors in \cite{Peng2021} introduced generative learning to their approach. This approach uses a 3D U-Net to encode spatio-temporal features. The output of the U-Net is both a reconstruction module and a classification module that predicts a pseudo-label for each sample using the reconstruction (similar to Figure \ref{fig:xdc}).

\subsubsection{Discussion}
In self-supervised contrastive learning, we discussed approaches that generate positive and negative samples via 1) temporal augmentation or 2) spatio-temporal augmentation. We also discussed approaches that 3) directly utilize the embedding space via clustering. 

When these approaches are evaluated on action recognition (shown in Table \ref{tab:action_recognition_contrastive} and Figure \ref{fig:ssl_contrastive}), spatio-temporal augmentation for generating samples had the highest success. The most successful approaches were those that were directly extended from image-based research to video by adding a temporal consistency \cite{Feichtenhofer_2021_CVPR}. This indicates that \textit{image-based methods may work just as well for video as long as temporal elements are added}. The clustering approach that performed the best was \cite{Alwassel2020} which used audio and video signal to generate pseudo-labels for the other. These results indicate that using additional signals may provide equally well performance without the need of data augmentation.

%% file: sections/crossmodal.tex
We have already discussed several works which utilize cross-modal signals in a hybrid setting, but this section will provide an comprehensive overview of approaches that solely use cross-modal agreement as their self-supervised learning objective. These works primarily use a contrastive NCE loss comparing positive and negative pairs.

\subsubsection{Video and Text}
The foundational work in this area pairs text and video \cite{Miech2019, Miech2020}. A new loss formulation was proposed in \cite{Miech2020} that is tailored to the challenges of using large-scale user-generated videos as found in the dataset HowTo100M \cite{Miech2019}, where 50\% of clip-narration pairs were misaligned. This loss, termed as \textit{Multiple Instance Learning Noise Contrastive Estimation} (MIL-NCE) loss, is a variation of the NCE loss and is shown in equation \ref{eq:mil_nce},
\begin{equation}
    \max_{f, g} \sum\limits^n_{i=1} \log \left( \frac{\sum\limits_{(x,y)\in \mathcal{P}_i}e^{f(x)^\top g(y)}}{\sum\limits_{(x,y)\in \mathcal{P}_i}e^{f(x)^\top g(y)} + \sum\limits_{(x',y')\sim \mathcal{N}_i}e^{f(x')^\top g(y')}} \right)
    \label{eq:mil_nce}
\end{equation}
where $(x,y)$ is a clip-narration pair and $f$, $g$ are visual and text encoders. The proposed method constructs a bag of positive candidate pairs $\mathcal{P}_i$ from the nearest captions in an initial time-range, then nearest captions when the initial-time range is tripled in duration. Negative samples are selected from the set of negative video/narration pairs $\mathcal{N}_i$. \cite{Miech2019} proposed a baseline for such an approach using instructional videos. More recently, VideoCLIP \cite{xu-etal-2021-videoclip} uses a contrastive loss of overlapping positive and negative text pairs by performing the reverse of \cite{Miech2019}'s strategy. The proposed method first samples a text snippet and then samples a video timestamp, within the boundaries of the clip, as a center. It then extended the video clip with a random duration from this center timestamp. In \cite{Amrani2020}, the authors propose to use a pairwise comparison where, in latent space, if two video clips are similar, then their captions should also be similar. A pairs' local \textit{k}-NN density estimation was used to measure the probability of a pair being correctly matched. The objective is formulated as a \textit{Soft Max Margin Ranking loss} \cite{wang2014learning, schroff2015facenet} where positive and negative pairs are from the same video/caption and different caption/video respectively, which are then weighted by the similarity weights. 

\begin{wraptable}{r}{.6\textwidth}
\caption{Downstream action segmentation evaluation on COIN for models that use a cross-modal agreement self-supervised pre-training approach. The top score is in \textbf{bold} and the second best is \underline{underlined}.}
\label{tab:action_segmentation_coin}
    \centering
    \resizebox{.59\textwidth}{!}{\begin{tabular}{lllll}
\toprule
& \multicolumn{2}{c}{Backbone} &           &                  \\
Model    &    Visual       &   Text        &                    Pre-train &                 Frame-Acc                    \\
\midrule
CBT \cite{Sun2019a}                          &     S3D-G &      BERT &  Kinetics+How2 &              53.90 \\
ActBERT \cite{Zhu2020}                       &   3D R-32 &      BERT &  Kinetics+How2 &              56.95 \\
VideoClip (zs) \cite{xu-etal-2021-videoclip} &     S3D-g &      BERT &           How2 &              58.90 \\
MIL-NCE \cite{Miech2020}                     &       S3D &  Word2Vec &           How2 &              61.00 \\
VLM \cite{Xu2020}                            &     S3D-g &      BERT &           How2 &              68.39 \\

VideoClip (ft) \cite{xu-etal-2021-videoclip} &     S3D-g &      BERT &           How2 &  \underline{68.70} \\
UniVL \cite{Luo2020}                         &     S3D-g &      BERT &           How2 &     \textbf{70.20} \\
\bottomrule
\end{tabular}}
\end{wraptable}

Some recent works, such as \cite{patrick2020support, Amrani2020}, extend the contrastive loss for HowTo100M pre-training by enforcing both global and local representations across samples. The authors in \cite{patrick2020support} proposed to replace the visual encoder with a transformer, as in \cite{Gabeur2020}, and used both a contrastive loss between video and text features as well as a generative loss using the entire batch. A reconstruction is generated as a weighted combination of video embeddings from a ``support-set'' of video clips selected via attention to enforce representation sharing between different samples.

The methods discussed so far learned independent encodings for each modality. In contrast, some of the approaches have attempted to learn joint-embedding for multiple signals \cite{Sun2019b, Sun2019a, liu2019use, Gabeur2020, miech2018learning, Wray2019}. In \cite{Wray2019}, the authors propose a cross-modal margin triplet loss where a video is paired with text and the text is also paired with a video. This method uniquely process text by separating nouns and verbs in two separate branches and pairs them separately with video using four different encoders. Then noun, verb and video representations are aggregated together via two encoders using the output from the other branches. The two final representations are then paired together for the final loss while negative pairs are generated from a mismatch between video and text. 

\begin{wraptable}{r}{.6\textwidth}
\caption{Downstream temporal action step localization
evaluation on CrossTask for models that use a contrastive multimodal self-supervised pre-training approach. Top scores are in \textbf{bold} and second best \underline{underlined}. }
\label{tab:action_localization_crosstask}
    \centering
    \resizebox{.59\textwidth}{!}{\begin{tabular}{lllll}
\toprule
& \multicolumn{2}{c}{Backbone} &           &                  \\
Model    &    Visual       &   Text        &                    Pre-train &                 Recall                    \\
\midrule
VideoClip (zs) \cite{xu-etal-2021-videoclip} &     S3D-g &      BERT &           How2 &              33.90 \\
MIL-NCE \cite{Miech2020}                     &       S3D &  Word2Vec &           How2 &              40.50 \\
ActBERT \cite{Zhu2020}                       &   3D R-32 &      BERT &  Kinetics+How2 &              41.40 \\
UniVL \cite{Luo2020}                         &     S3D-g &      BERT &           How2 &              42.00 \\
VLM \cite{Xu2020}                            &     S3D-g &      BERT &           How2 &  \underline{46.50} \\
VideoClip (ft) \cite{xu-etal-2021-videoclip} &     S3D-g &      BERT &           How2 &     \textbf{47.30} \\

\bottomrule
\end{tabular}}
\end{wraptable}
Rather than relying on separate embeddings, the authors in \cite{Sun2019a} concatenated text and video embeddings together before passing them to a cross-modal transformer. The joint-embeddings are formulated to generate positive and negative pairs of aligned and unaligned ASR and video to measure the NCE loss. A more complex approach  in \cite{Gabeur2020} extended prior work \cite{liu2019use, miech2018learning}. This model learns to maximize the similarity between the associated text of a video and a cross-modal embedding for video using ``experts'' \cite{liu2019use} and a complex aggregation procedure. Each expert encodes either motion, audio, scene, OCR, face, speech and appearance of video clips which are concatenated together. To know which of these features came from which encoder, it generates a mapping vector. The temporal representations indicates the time in the video where each feature is extracted. These three vectors are then aggregated together and passed to a transformer to output the final representation. Text is embedded using BERT and gated embedding modules \cite{miech2018learning}, one for each expert on the video embedding side. 

\begin{table}[]
    \centering
    \caption{Performance for the downstream video retrieval task. Top scores for each category are in \textbf{bold} and second best \underline{underlined}. Masked Modeling (MM) is a generative approach that uses both video with text. Cross-modal agreement include a variety of contrastive approaches that can use video with audio and/or text. Cross-modal agreement pre-training approaches typically perform best. Some models have dedicated variations in what they report with fine-tuning (*) on the target dataset, YouCook2 or MSRVTT. The pre-training datasets titled COMBO are CC3M \cite{sharma-etal-2018-conceptual}, WV-2M\cite{Bain21} and COCO \cite{coco_dataset}.}
    \resizebox{\textwidth}{!}{\begin{tabular}{lllllll|lll}
\toprule
 &       \multicolumn{2}{c}{ Backbone} &       &    \multicolumn{3}{c}{YouCook2}  &          \multicolumn{3}{c}{MSRVTT} \\
 Model     &      Visual  & Text            &   Pre-Train                         & R@1                   &       R@5             &     R@10               &        R@1         &          R@5          &        R@10            \\

\hline
\multicolumn{10}{c}{Masked Modeling}\\
\hline
ActBERT \cite{Zhu2020}                              &       3D R-32 &          BERT &  Kinetics+How2 &               9.60 &              26.70 &              38.00 &               8.60 &              23.40 &              33.10 \\
HERO \cite{Li2020}                                &      SlowFast &    WordPieces &        How2+TV\cite{tv_dataset} &                -- &                -- &                -- &              16.80 &              43.40 &              57.70 \\
ClipBERT \cite{Lei2021}                            &          R-50 &    WordPieces &        VisualGenome \cite{krishnavisualgenome} &                -- &                -- &                -- &              \underline{22.00} &              46.80 &              59.90 \\

VLM \cite{Xu2020}                                 &         S3D-g &          BERT &           How2 &              \underline{27.05} &              \underline{56.88} &              \underline{69.38} &              \textbf{28.10} &              \textbf{55.50} &             \textbf{67.40} \\
UniVL \cite{Luo2020}                             &         S3D-g &          BERT &           How2 &              \textbf{28.90} &              \textbf{57.60} &              \textbf{70.00} &              21.20 &              \underline{49.60} &              \underline{63.10} \\

\hline
\multicolumn{10}{c}{Cross-Modal Agreement}\\
\hline
Amrani et al.  \cite{Amrani2020}                &         R-152 &      Word2Vec &           How2 &                -- &                -- &                -- &               8.00 &              21.30 &              29.30 \\
MIL-NCE \cite{Miech2020}                           &           S3D &      Word2Vec &           How2 &              15.10 &              38.00 &              51.20 &               9.90 &              24.00 &              32.40 \\
COOT \cite{COOT}                                   &         S3D-g &          BERT &  How2+YouCook2 &              16.70 &              40.20 &              52.30 &                 -- &                 -- &                 -- \\
CE*  \cite{liu2019use}                                &       Experts &       NetVLAD \cite{Arandjelovic_2016_CVPR} &                MSRVTT &                -- &                -- &                -- &              10.00 &              29.00 &              41.20 \\
VideoClip  \cite{xu-etal-2021-videoclip}        &         S3D-g &          BERT &           How2 &              22.70 &              50.40 &              63.10 &              10.40 &              22.20 &              30.00 \\
VATT \cite{akbari2021vatt}                          &  Linear Proj. &  Linear Proj. &  AS+How2 &                -- &                -- &              40.60 &                -- &                -- &              23.60 \\

MEE \cite{miech2018learning}                        &       Experts &       NetVLAD \cite{Arandjelovic_2016_CVPR} &           COCO\cite{coco_dataset} &                -- &                -- &                -- &              14.20 &              39.20 &              53.80 \\
JPoSE \cite{Wray2019}                               &           TSN \cite{wang2016temporal} &      Word2Vec &            Kinetics &                -- &                -- &                -- &              14.30 &              38.10 &              53.00 \\

Amrani et al.*  \cite{Amrani2020}                &         R-152 &      Word2Vec &           How2 &                -- &                -- &                -- &              17.40 &              41.60 &              53.60 \\
AVLnet* \cite{Rouditchenko2020}                      &      3D R-101 &      Word2Vec &    How2	 &                \underline{30.20} &              \underline{55.50} &              \underline{66.50} &              22.50 &              50.50 &              64.10 \\
MMT \cite{Gabeur2020}                          &       Experts &          BERT &           How2 &                -- &                -- &                -- &                -- &              14.40 &                -- \\
MMT*  \cite{Gabeur2020}                          &       Experts &          BERT &           How2 &                -- &                -- &                -- &                -- &  55.70 &                -- \\

Patrick et al.*  \cite{patrick2020support}       &       Experts &           T-5 \cite{raffel2019exploring} &           How2 &                -- &                -- &                -- &  30.10 &     \underline{58.50} &     \underline{69.30} \\
VideoClip* \cite{xu-etal-2021-videoclip}        &         S3D-g &          BERT &           How2 &  \textbf{33.20} &  \textbf{62.60} &  \textbf{75.50} &     \underline{30.90} &              55.40 &              66.80 \\

FIT \cite{Bain21} & ViT \cite{dosovitskiy2020image} & BERT & COMBO & -- & -- & -- & \textbf{32.50} & \textbf{61.50} & \textbf{71.20} \\

\bottomrule
\end{tabular}}
    
    \label{tab:text_video_retrieval_all}
\end{table}

\subsubsection{Video and Audio}

Visual representation learning using both audio and video has been proposed in several works \cite{8682475, Morgado2020, Korbar2018}. Using basic data augmentation, multiple types of transformations that were hierarchically sampled and applied to each batch were used in  \cite{Patrick2020}. This generated distinct and/or invariant positive pairs. These transformations are temporal, cross-modal, and spatial/aural augmentations. The cross-modal transformation projected the video to either its visual or audio component.

Temporal augmentation has been used for contrastive synchronization tasks between audio and video modalities  \cite{Morgado2020,Rouditchenko2020,Owens2018,Afouras2020,Korbar2018}. Traditional approaches used a two branch network, one for audio and another for video, followed by a fusing procedure \cite{Korbar2018, Owens2018}. The authors in \cite{Owens2018} used a binary classification task between video and audio that is synchronized from the same video, or video with audio that was shifted several seconds. Meanwhile in \cite{Korbar2018}, a contrastive loss used traditionally for Siamese networks \cite{koch2015siamese} was used with \textit{easy negatives} and \textit{hard negatives}. \textit{Easy negatives} were frames and audio from two different videos whilst \textit{hard negatives} were frames and audio from the same video with short time-gaps positioned between them.
In recent works, \cite{Morgado2020} uses an \textit{audio-visual correspondence task}, first proposed in \cite{Arandjelovic_2018_ECCV}, by taking an exponential moving average of representations, called \textit{memory representations}, for video and audio features that are maintained to make contrastive and cross-modal comparisons. The probability that a \textit{memory representation} for a sample, $\overline{v}_i$ or $\overline{a}_i$, belongs to a given sample $v_i$ or $a_i$, is modeled as a softmax function. Negative samples are a number of instances, drawn from the distribution at random, from different videos. Whereas positive samples were the memory representations and their corresponding samples. The authors use a cross-modal NCE contrastive loss and an cross-modal agreement loss. The cross-modal agreement use positive samples that are the top-k most similar to the anchor sample in both spaces while negative are the top-k most distant. 

Clustering pseudo-labels has also been used with audio and video. 
Inspired by DeepCluster for images \cite{caron2018deep}, \cite{Alwassel2020} uses audio signals to generate pseudo-labels for video and vice versa. This is achieved using a classification task as opposed to contrastive learning as shown in Figure \ref{fig:xdc}. This approach uses a video encoder and an audio encoder to embed RGB frames and audio spectrograms respectively. In each iteration, the features are clustered separately. Video uses cluster assignments from audio while audio uses cluster assignments from video representations. The authors in \cite{Asano2020} also utilize clustering but use information from both audio and video during each iteration. The clustering optimization is formulated as an optimal transport problem with a tail-end prior distribution for clustering assignments. This method extracts a sample's average latent representations from encoders using modality splicing and data augmentations. The activations that are used for clustering are averaged over these augmentations. This results in clusters that are invariant to augmentations and choice of modality. 

\subsubsection{Video, Text and Audio}
Some approaches utilize signals from audio, visual and text associated with a video. Using the three signals was first proposed in the image domain \cite{Kaiser2017,Aytar2017} for tasks such as cross-modal retrieval and the transferring of classifiers between modalities. 
In \cite{Rouditchenko2020}, features are extracted using encoders designed specifically for each signal. A non-linear feature gating is then used to generate a joint-embedding between the three. Using a Masked Margin Softmax Loss \cite{ilharco2019large,oord2018representation}, comparisons were made between audio-visual pairs, visual-text pairs, and audio-text pairs. Clustering for pseudo-labels has also been used similar to \cite{Alwassel2020} and Figure \ref{fig:xdc}. The authors in \cite{Chen2021} added text as one of the signals with this approach. More specifically, the authors aggregated text, audio, and visual features into a joint space and then performed clustering. The overall objective used three tasks: a contrastive loss, a clustering loss, and a reconstruction loss simultaneously.

To account for the differences in video, audio, and text modalities, \cite{Alayrac2020} proposes to first joint-embed video and audio because they are at a fine-grained granularity. Then the joint-embedding is embedded with text, which is coarse-grained. This architecture uses a two-branch embedding of audio and visual features separately. Training uses a traditional NCE loss between the joint audio-visual embeddings and an MIL-NCE loss \cite{Miech2019} between the joint visual-audio-text embeddings.
This common space projection method is later used in \cite{akbari2021vatt} as a Multimodal Projection Head. This approach discretized video, audio and text signal before passing to a traditional transformer. To reduce computational complexity, the authors propose DropToken, randomly dropping these tokens before passing to a transformer. The Multimodal Projection Head uses the output of the transformer for final shared space projection and contrastive learning.

\begin{table}
\caption{Downstream action recognition on UCF101 and HMDB51 for models that use contrastive learning and/or cross-modal agreement. Top scores for each category are in \textbf{bold} and second best are \underline{underlined}. Modalities include video (V), optical flow (F), human keypoints (K), text (T) and audio (A). Spatio-temporal augmentations with contrastive learning typically are the highest performing approaches. 
}
\label{tab:action_recognition_contrastive}
    \centering
    \resizebox{\textwidth}{!}{
\begin{tabular}{lllllllll}
\toprule
 &        &    \multicolumn{3}{c}{Backbone} &  & & \multicolumn{2}{c}{Accuracy} \\
 \cline{3-5} \cline{8-9}
Model     &     Subcategory &    Visual    &    Text      &    Audio       &    Modalities        &      Pre-Train               &      UCF101              &HMDB51\\
\midrule
VIE \cite{Zhuang_2020_CVPR}                     &         Clustering &     Slowfast &           -- &        -- &          V &            Kinetics &              78.90 &               50.1 \\
VIE-2pathway \cite{Zhuang_2020_CVPR}            &         Clustering &         R-18 &           -- &        -- &          V &            Kinetics &  \underline{80.40} &               52.5 \\
Tokmakov et al. \cite{tokmakov2020unsupervised} &         Clustering &      3D R-18 &           -- &        -- &          V &            Kinetics &     \textbf{83.00} &               50.4 \\

\hline
TCE \cite{Knights2020}                          &           Temporal Aug.  &         R-50 &           -- &        -- &          V &              UCF101 &  \underline{71.20} &               36.6 \\
Lorre et al. \cite{Lorre2020}                   &            Temporal Aug. &         R-18 &           -- &        -- &        V+F &              UCF101 &     \textbf{87.90} &               55.4 \\
\hline
CMC-Dual \cite{Tian2020_CMC}                    &           Spatial Aug. &     CaffeNet &           -- &        -- &        V+F &              UCF101 &              59.10 &               26.7 \\
SwAV \cite{Feichtenhofer_2021_CVPR}             &           Spatial Aug. &         R-50 &           -- &        -- &          V &            Kinetics &              74.70 &                 -- \\
VDIM \cite{Hjelm2020}                           &           Spatial Aug. &      R(2+1)D &           -- &        -- &          V &            Kinetics &              79.70 &               49.2 \\
CoCon \cite{Rai_2021_CVPR}                      &           Spatial Aug. &         R-34 &           -- &        -- &      V+F+K &              UCF101 &              82.40 &               53.1 \\
SimCLR \cite{Feichtenhofer_2021_CVPR}           &           Spatial Aug. &         R-50 &           -- &        -- &          V &            Kinetics &              84.20 &                 -- \\
CoCLR \cite{Han2020_cotraining}                 &           Spatial Aug. &        S3D-G &           -- &        -- &        V+F &              UCF101 &              90.60 &               62.9 \\
MoCo \cite{Feichtenhofer_2021_CVPR}             &           Spatial Aug. &         R-50 &           -- &        -- &          V &            Kinetics &  \underline{90.80} &                 -- \\
BYOL \cite{Feichtenhofer_2021_CVPR}             &           Spatial Aug. &         R-50 &           -- &        -- &          V &            Kinetics &     \textbf{91.20} &                 -- \\

\hline

MIL-NCE \cite{Miech2020}                        &         Cross-Modal &        S3D-G &     Word2Vec &        -- &        V+T &           How2 &              61.00 &               91.3 \\
GDT \cite{Patrick2020}                          &         Cross-Modal &      R(2+1)D &     Word2Vec &    2D CNN &      V+T+A &            Kinetics &              72.80 &               95.5 \\
CBT \cite{Sun2019a}                             &         Cross-Modal &        S3D-G &         BERT &        -- &        V+T &            Kinetics &              79.50 &               44.6 \\
VATT \cite{akbari2021vatt}                      &         Cross-Modal &  Transformer &  Transformer &        -- &        V+T &  AS+How2 &              85.50 &               64.8 \\
AVTS \cite{Korbar2018}                          &         Cross-Modal &          MC3 \cite{tran2018closer} &           -- &       VGG &        V+A &            Kinetics &              85.80 &               56.9 \\
AVID+Cross \cite{Morgado2020}                   &         Cross-Modal &      R(2+1)D &           -- &    2D CNN &        V+A &            Kinetics &              91.00 &               64.1 \\
AVID+CMA \cite{Morgado2020}                     &         Cross-Modal &      R(2+1)D &           -- &    2D CNN &        V+A &            Kinetics &              91.50 &               64.7 \\
MMV-FAC \cite{Alayrac2020}                      &         Cross-Modal &          TSM \cite{lin2019tsm} &     Word2Vec &      R-50 &      V+T+A &  AS+How2 &  \underline{91.80} &               67.1 \\
XDC \cite{Alwassel2020}                         &         Cross-Modal &      R(2+1)D &           -- &    ResNet &        V+A &            Kinetics &     \textbf{95.50} &               68.9 \\
\hline

DVIM \cite{Xue2020}                             &  Spatio-Temporal Aug. &         R-18 &           -- &        -- &        V+F &              UCF101 &              64.00 &               29.7 \\
IIC \cite{Tao2020}                              &  Spatio-Temporal Aug. &          R3D &           -- &        -- &        V+F &            Kinetics &              74.40 &               38.3 \\
DSM \cite{Wang2020}                             &  Spatio-Temporal Aug. &          I3D &           -- &        -- &          V &            Kinetics &              78.20 &               52.8 \\
pSimCLR \cite{chen2020simple}                   &  Spatio-Temporal Aug. &         R-50 &           -- &        -- &          V &            Kinetics &              87.90 &                 -- \\
TCLR \cite{DAVE2022103406}                      &  Spatio-Temporal Aug. &      R(2+1)D &           -- &        -- &          V &              UCF101 &              88.20 &               60.0 \\
SeCo \cite{Yao2020}                             &  Spatio-Temporal Aug. &         R-50 &           -- &        -- &          V &            ImageNet\cite{NIPS2012_c399862d} &              88.30 &               55.6 \\
pSwaV \cite{caron2021unsupervised}              &  Spatio-Temporal Aug. &         R-50 &           -- &        -- &          V &            Kinetics &              89.40 &                 -- \\
pBYOL \cite{grill2020bootstrap}                 &  Spatio-Temporal Aug. &         R-50 &           -- &        -- &          V &            Kinetics &              93.80 &                 -- \\
CVRL \cite{Qian2020}                            &  Spatio-Temporal Aug. &      3D R-50 &           -- &        -- &          V &            Kinetics &              93.90 &               69.9 \\
pMoCo \cite{he2020momentum}                     &  Spatio-Temporal Aug. &         R-50 &           -- &        -- &          V &            Kinetics &  \underline{94.40} &                 -- \\
pBYOL \cite{Feichtenhofer_2021_CVPR}            &  Spatio-Temporal Aug. &        S3D-G &           -- &        -- &          V &            Kinetics &     \textbf{96.30} &               75.0 \\
\bottomrule
\end{tabular}}
\end{table}

\subsubsection{Discussion}
In cross-modal learning, we discussed generating positive and negative samples via 1) video and audio signals 2) video and text signals 3) and video, audio and text signals. These multimodal approaches typically pre-train on the HowTo100M dataset \cite{Miech2019}. Many of these models, such as UniVL \cite{Luo2020}, VideoClip \cite{xu-etal-2021-videoclip} and COOT \cite{COOT}, rely on pre-extracting features from the S3D-g visual encoder from \cite{Miech2020}.  \textit{Future work should attempt to understand more what these models are capable of learning by using raw video as input}, especially with transformer based architectures like VATT \cite{akbari2021vatt}. 

\begin{wrapfigure}{R}{0.5\textwidth}
    \includegraphics[width=.5\textwidth]{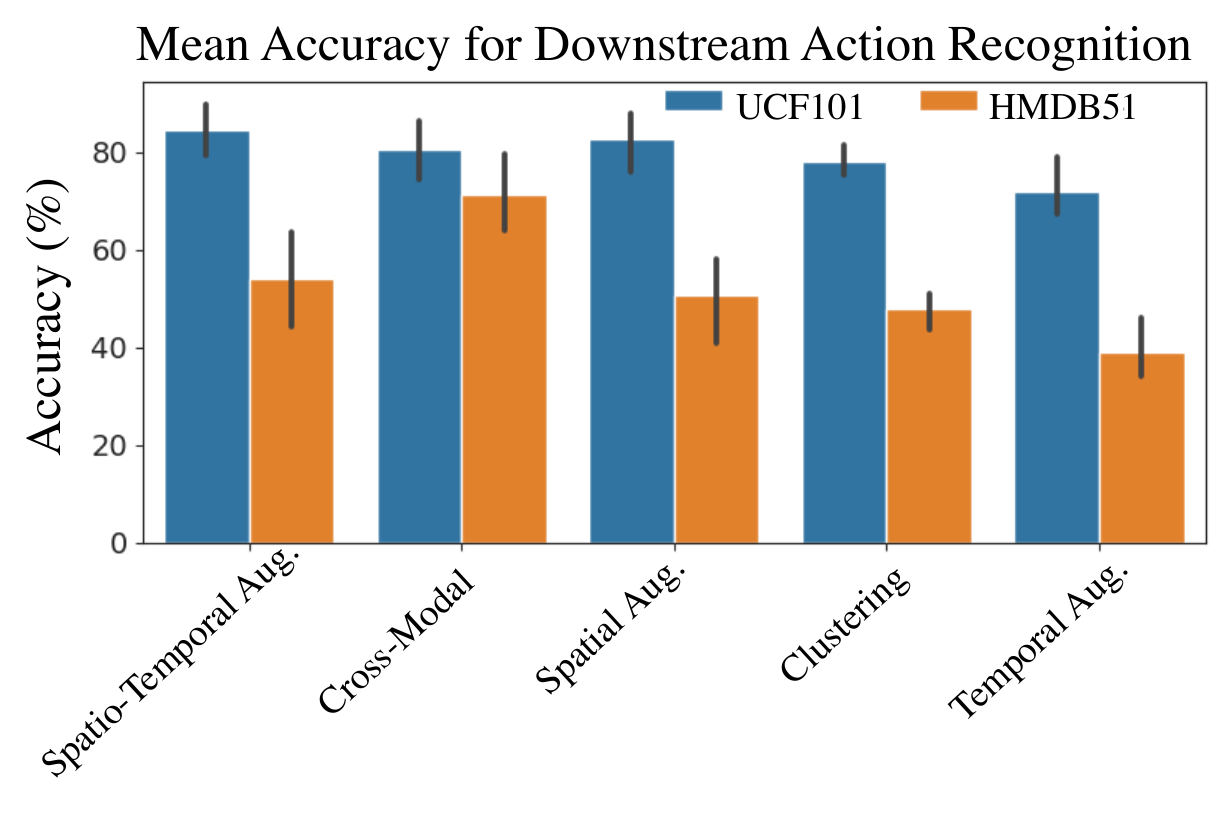}
    \caption{Mean performance on action recognition for approaches mentioned in Table \ref{tab:action_recognition_contrastive}. Using spatio-temporal augmentations with contrastive learning typically perform best.
    }
    \label{fig:ssl_contrastive}
\end{wrapfigure}

When these approaches are evaluated on action recognition (shown in Table \ref{tab:action_recognition_contrastive} and Figure \ref{fig:ssl_contrastive}), the multimodal approach that performed the best was \cite{Alwassel2020}, which used audio and video signal to generate psuedo-labels for the other, using a classification task as opposed to contrastive learning. The other multimodal approach that performed well was \cite{Alayrac2020}, which used a contrastive loss between embeddings of text, audio and video, using joint-modal embeddings. 
Multimodal approaches also generalize to more downstream tasks as shown in Table \ref{tab:action_segmentation_coin}, \ref{tab:text_video_retrieval_all}, and \ref{tab:action_localization_crosstask}. In Table \ref{tab:text_video_retrieval_all}, multimodal contrastive losses outperform masked modelling (MM). Models that are fine-tuned on the target dataset, as expected, perform the best as seen with VideoClip \cite{xu-etal-2021-videoclip}. This may indicate that \textit{cross-modal learning shows better performance than joint-modal learning for some tasks} like text-to-video retrieval and temporal action step localization. 


%% file: sections/summary_future_work.tex
 
This survey discussed pretext, generative, contrastive and multimodal approaches to self-supervised video representation learning. 
\textit{Pretext tasks} that are most commonly used includes frame order, appearance statistics, jigsaw, and playback speed related tasks. Of the pretext learning tasks, classifying playback speed performs the best on downstream action recognition and is one of the more recent video only based approaches. This approach performs better than frame ordering because it has multiple ways to use temporal signal by modifying the speed while frame order has limited permutations. \textit{Generative approaches} commonly used include generative adversarial networks, masked auto-encoders and multimodal masked modeling. Some of these approaches used a sequence of previous frames to generate the next sequence of frames. Other approaches used a model to generate masked input using the unmasked input of one or multiple modalities.

The approaches that performed the best used masked modeling. \textit{Contrastive approaches} are the most common approach to self-supervised learning of video. There are many ways positive and negative samples are generated for an anchor clip including view augmentation, temporal augmentation, spatio-temporal augmentation, clustering and multimodal. Contrastive learning is the best performing learning objective in all action based downstream tasks. These approaches have also entered a multimodal domain with \textit{cross-modal agreement}. These approaches both improve performance in action-recognition tasks and expand into other tasks like text-to-video retrieval and action segmentation using text and audio modalities.


\subsection{Limitations and Future Work}
\begin{figure}
    \centering
    \includegraphics[width=.95\textwidth]{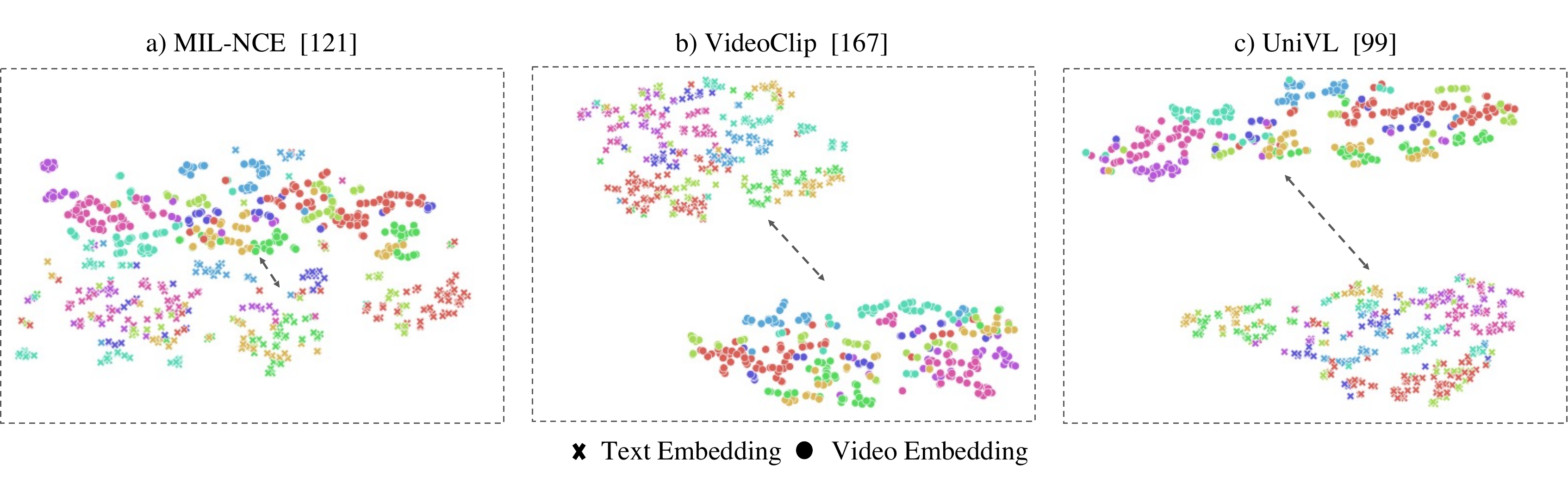}
    \caption{T-sne visualizations of multimodal features where video and text are in a joint space. The colors indicate the recipe of clips from the YouCook2 dataset. The shapes of each point indicate either text or video embeddings. 
    When visualized, even though the VideoClip \cite{xu-etal-2021-videoclip} and UniVL \cite{Luo2020} models significantly outperform the MIL-NCE model \cite{Miech2020} on downstream tasks, the text and visual embeddings show a large gap in the latent space compared to to MIL-NCE. Future work should look at why these higher performing models produce multimodal embeddings that look distant from each other.}
    \label{fig:multimodal_embedding_space}
\end{figure}

While the existing approaches improved the generalizability of learned representations, there are still some limitations which need to be addressed. For example, when evaluating on downstream action recognition, the majority of approaches use the UCF101 \cite{soomro2012ucf101} and HMDB51 \cite{kuehne2011hmdb} datasets that lack temporal specificity. More approaches should additionally evaluate on temporal specific datasets like the Something-Something dataset \cite{goyal2017something}. 
Specific to the different categories, future work should attempt to understand the performance and robustness between cross-modal approaches. 
Moreover, detection tasks such as video object detection, action detection, video semantic segmentation, etc. are not well explored from self-supervision point of view and are promising future directions. Additionally, combining learning objectives from multiple categories, such as masked modeling and contrastive learning, can be be further explored as preliminary works already show increased performance \cite{chung2021w2v,Luo2020,COOT}.

\textbf{\textit{Long-term video representation}}.
Current approaches mostly focus on small video clips and do not learn representations for a long video.  In most approaches, a video is split into clips as part of their pre-processing. While these approaches improve the learning using temporal elements of the video, they lack long-term temporal learning. Furthermore, it has been demonstrated that datasets like MSRVTT and UCF101 do not necessarily need temporal learning \cite{huang2018makes,buch2022revisiting}. Future research should evaluate on more temporally relevant datasets like Something-Something \cite{goyal2017something} and NExT-QA \cite{xiao2021next}. Future research should also attempt to \textit{learn a representation from a long video as a whole, focusing on temporal learning rather than only using the temporal signal as a guidance for training. }

\textbf{\textit{Simple end-to-end approaches}}.
Most of the approaches use a pre-trained backbone as a dependency for feature extraction. For example, many of the multimodal approaches use the S3Dg from MIL-NCE \cite{Miech2019}. This model is already pre-trained using a contrastive loss function on text and video and is a large, computational approach out-of-the-box. These new approaches extract features from the S3Dg, then add-on large, computational approaches. There has been little focus on reducing the size of these models or to create an approach that does not rely on previous works directly. Additionally, this prevents probing models for better understanding of what these models are learning, especially when there is semantic information available via text. This is also the same for video-only approaches using networks like I3D \cite{carreira2018quo}, when the downstream task is also action recognition. There are recent approaches that utilize a transformer visual encoder, such as \cite{akbari2021vatt} and \cite{ranasinghe2022self}, that takes raw video as input, but there is limited research in this area but great potential for future works. \textit{Future work should focus more on simple end-to-end approaches that do not rely on previous work so directly and therefore can reduce the computational costs and dependencies.} 

\textbf{\textit{Interpretability and Explainability}}.
Downstream tasks have focused on specific areas to each modality, such as action recognition, event recognition for audio, and caption generation for text. While these help understand the generalizability of the learned representations, they do not provide insight to what these models are actually learning. While this has been done in the image-domain by probing models \cite{thrush2022winoground}, there is no approach that demonstrates the understanding a model uses to generate the representations in the video domain. This lack of interpretability enforces the `black-box model" approach. \textit{Future research should focus on understanding what these models are learning and why their representations are more generalizable}. With this understanding, models can be further improved and provide more understanding about videos.
While multimodal approaches typically are the best performers for downstream tasks, the lack of insight on what the models are learning prevents an understanding of how these modalities are interacting. For example, when using two modalities, what does the resulting joint-embedding space look like? In Figure \ref{fig:multimodal_embedding_space}, an example of this joint space is visualized using t-sne. The samples are clips from different recipes in the YouCook2 dataset, where visual features are depicted as $\boldsymbol{\cdot}$ and text features are $x$. While the cross-modal VideoClip \cite{xu-etal-2021-videoclip} and joint-modal UniVL \cite{Luo2020} models significantly outperform the MIL-NCE model \cite{Miech2020} on the text-to-video retrieval task, the feature embeddings for each modality are distant from each other. Future work should investigate questions like this. More interpretable models would allow better representation of how text, video and audio are interacting and could lead to improved semantic graphs, temporal understanding of event interactions, and more. 

\textbf{\textit{Standard Benchmarking}}.
The variation in backbone encoders and pre-training protocols makes direct comparisons difficult. While this survey attempts to organize such differences and discusses several existing benchmarks \cite{Feichtenhofer_2021_CVPR,huang2018makes,buch2022revisiting}, it would be beneficial to the community to have a benchmark evaluation for self-supervised learning that allows better comparisons between the many different approaches. While in the image domain, robustness has been investigated for self-supervised learning \cite{hendrycks2018benchmarking,bhojanapalli2021understanding,hendrycks2021many}, there is a shortage of similar work in the video domain \cite{robustness_multimodal,liang2021multibench}. These models are often evaluated on zero-shot learning, so understanding the robustness to different types of distribution shifts will provide further improvements. \textit{Future work should generate benchmarks for evaluation on the performance of self-supervised learning as well as the robustness of different approaches.}



\section{Conclusion}
\label{sec:conclusion}
This survey discussed self-supervised learning for video and analyzed the current landscape. From our analysis, we observed the following interesting points; 1) playback speed or temporal consistency, typically performed better for pretext tasks, 2) image-based augmentation methods for contrastive learning work just as well for video, as long as temporal elements are included 3), and multimodal approaches can provide equally as well performance to unimodal approaches without the need of data augmentations.
This survey also discussed several limitations in existing works. These limitations include: 1) lack of standardized benchmarks for performance, 2) heavy reliance on existing visual encoders which operates on trimmed video clips, 3) and a lack of understanding on what these models are actually learning and what makes them more generalizable. From the limitations discussed, we put forward areas for future work. These include: 1) long-term video representation, 2) simple end-to-end approaches, 3) interpratibility and explainability, 4) and standard benchmarking.  We believe this survey will serve as an important resource for the research community and guide future research in self-supervised video representation learning.